A. Mantelero, M.S. Esposito[*]

**An evidence-based methodology for human rights impact assessment (HRIA) in the development of AI data-intensive systems.**

A revised version of this paper is published in Computer Law & Security Review, Volume 41, July 2021, 105561, available here in open access:

https://www.sciencedirect.com/science/article/pii/S0267364921000340#cit_12

*Different approaches have been adopted in addressing the challenges of Artificial Intelligence (AI), some centred on personal data and others on ethics, respectively narrowing and broadening the scope of AI regulation. This contribution aims to demonstrate that a third way is possible, starting from the acknowledgement of the role that human rights can play in regulating the impact of data-intensive systems.*

*The focus on human rights is neither a paradigm shift nor a mere theoretical exercise. Through the analysis of more than 700 decisions and documents of the data protection authorities of six countries, we show that human rights already underpin the decisions in the field of data use.*

*Based on empirical analysis of this evidence, this work presents a methodology and an assessment model for a Human Rights Impact Assessment (HRIA). The methodology and related assessment models are focused on AI applications, whose nature and scale require a proper contextualisation of HRIA methodology. Moreover, the proposed models provide a more measurable approach to risk assessment which is consistent with the regulatory proposals centred on risk thresholds.*

*The proposed methodology is tested in concrete case-studies to prove its feasibility and effectiveness. The overall goal is to respond to the growing interest in HRIA, moving from a mere theoretical debate to a concrete and context-specific implementation in the field of data-intensive applications based on AI.*

**1. Introduction**

The debate that has characterised the last few years on data and Artificial Intelligence (AI) represents an interesting arena in which to consider the theoretical evolution of the future approach in addressing the challenges posed by AI to human rights. This debate has been marked by an emphasis on the ethical dimension of the use of algorithms[1] and, in the legal domain, by a focus on potential bias and protection from discrimination.[2]

---

[*] A. Mantelero coordinated this study and is author of all sections except 4.2. M.S. Esposito is author of section 4.2 and carried out the research described in that section. Corresponding author: Alessandro Mantelero, Associate Professor at Polytechnic University of Turin, alessandro.mantelero@polito.it. M.S. Esposito is Adjunct Professor at Polytechnic University of Turin. The authors would like to acknowledge the helpful feedback on the model design and multicriteria analysis received from Maria Franca Norese, Associate Professor of Operations Research at the Polytechnic University of Turin.

However, the emerging AI-driven society shows a variety of potential impacts on individual and collective rights and freedoms suggesting, on the one hand, a reaffirmation of the central role of the legal instruments, not replaceable by ethical guidelines, and, on the other hand, the need for a more comprehensive analysis of the rights and freedoms concerned.

This contribution, after some considerations on the interplay between ethical and legal approaches in the debate on AI regulation, focuses on the analysis of the impact of AI on human rights and the development and application of an AI-specific impact assessment model.

Empirical evidence on how data-intensive systems may affect rights and freedoms is drawn from the European context which is characterised by a long-standing and persistent focus on human rights – partly through the role of the European Court of Human Rights – and a theoretical approach that has intertwined data processing and human rights since the early data protection regulations.[3]

The 'case study' of Europe, which provides extensive jurisprudence on data processing developed by the data protection authorities, is used to achieve a broader global perspective in dealing with AI and human rights. Creating a methodological approach to impact assessment built on international legal instruments, we provide a model that can be more easily used in different legal cultures, grounded on the universal nature of the rights and freedoms in question.[4]

---

[1] See Luciano Floridi et al., 'AI4People—An Ethical Framework for a Good AI Society: Opportunities, Risks, Principles, and Recommendations' (2018) Minds & Machine, DOI: 10.1007/s11023-018-9482-5 accessed 30 November 2020; Brent Daniel Mittelstadt et al., 'The ethics of algorithms: Mapping the debate' (2016) Big Data & Society 1–21.

[2] See Sandra Wachter, Brent Mittelstadt, and Chris Russell, *Bias Preservation in Machine Learning: The Legality of Fairness Metrics Under EU Non-Discrimination Law* (2021) West Virginia Law Review, Forthcoming <https://papers.ssrn.com/abstract=3792772> accessed 27 February 2021; Algorithm Watch, 'Automating Society report 2020', <https://automatingsociety.algorithmwatch.org/wp-content/uploads/2020/12/Automating-Society-Report-2020.pdf> accessed 23 January 2021); Sarah Myers West, Meredith Whittaker & Kate Crawford, 'Discriminating Systems' 33 (2019), <https://ainowinstitute.org/discriminatingsystems.pdf> accessed 13 June 2020; Frederik J. Zuiderveen Borgesius, 'Strengthening legal protection against discrimination by algorithms and artificial intelligence' (2020) 24(10) Int. J. Hum. Rights 1572-1593; Monique Mann and Tobias Matzner, Challenging algorithmic profiling: The limits of data protection and anti-discrimination in responding to emergent discrimination, (2019) 6(2) Big Data & Society, DOI: 10.1177/2053951719895805.

[3] See e.g. Lee Andrew Bygrave, *Data Privacy Law: An International Perspective* (Oxford University Press, 2014); Gloria Gonzalez Fuster, *The Emergence of Personal Data Protection as a Fundamental Right of the EU* (Springer 2014); Viktor Mayer-Schönberger, Generational Development of Data Protection in Europe, in Philip E. Agre and Marc Rotenberg (eds) *Technology and Privacy: The New Landscape* (The MIT Press 1997) 219-241.

[4] Referring to this universal nature, we are aware of the underlying tensions that characterise it, the process of contextualisation of these rights and freedoms (appropriation, colonisation, vernacularisation, etc.) and the theoretical debate on universalism and cultural relativism in human rights. See Peggy Levitt and Sally Merry, 'Vernacularization on the Ground: Local Uses of Global Women's Rights in Peru, China, India and the United States' (2009) 9 Global Networks 441-461; Seyla Benhabib, 'The Legitimacy of Human Rights' (2008) 137 Daedalus 94-104; Sally Engle Merry, *Human rights and gender violence: translating international law into local justice* (Univ of Chicago Press 2006). See also Daniel M Goldstein, 'Human Rights as Culprit, Human Rights as Victim: Rights and Security in the State of Exception' in Mark Goodale and Sally Engle Merry (eds), *The Practice of Human Rights: Tracking Law between the Global and the Local* (Cambridge University Press 2007) 49-77; Lauren Leve, '"Secularism Is a Human Right!": Double-Binds of Buddhism, Democracy, and Identity in Nepal' ibid 78-114; Thomas Risse and Stephen C Ropp, 'International Human Rights Norms and Domestic Change: Conclusions' in Kathryn Sikkink, Stephen C Ropp and Thomas Risse (eds), *The Power of Human Rights: International*

The next two sections of this work highlight the need to turn our gaze to law after the focus on data ethics in recent years, suggesting a complementary role for ethics and law and stressing that only human rights can provide a uniform reference for regulating AI in different cultural, ethical and legal contexts.

Section 4 investigates how human rights can be impacted by data intensive systems, adopting an empirical evidence-based approach rather than a theoretical one. Instead of creating fictitious cases on human rights and AI, we conduct an empirical analysis of decided cases, identifying the interplay between the use of data and human rights. The European context was chosen as the case study, given the existence of a general regulation on data protection – not only the GDPR but also in the preceding decades, at national level and, after 1995, at EU level – with the availability of extensive case law and related records of practice.

This section does not take a European-centred perspective in addressing AI issues, but provides empirical evidence on situations where a clash between human rights and data intensive systems occurs. The goal is not to analyse the response of the European legal framework but to extract a list of potentially impacted rights and freedoms based on concrete evidence as opposed to theoretical cases and hypothesis. This evidence-based approach underpins the methodology of our analysis and the proposed assessment model.

This exercise also brings out the limitations of data protection regulations in addressing AI issues and the need for a more tailored approach, rather than broad notions such as fairness, to enlarge the scope of data protection regulation and encompass AI application and related matters.

Having defined the theoretical framework and gathered the empirical evidence, Sections 5 and 6 present the key product of the research, the development and testing of an AI-focused human rights impact assessment model. The structure and components of the model are described in Section 5, while Section 6 discusses its concrete application in two cases with very different impacts in terms of scale. The cases chosen, one global and another limited to Canada, refer to two different types of AI applications, a smart device and a smart city plan. Since both projects have now been concluded, contrafactual analysis of the available documentation was able to test the model and show its results and effects in those contexts.

The last section provides some concluding remarks and points out the potential benefit of adopting the proposed assessment model in terms of legal compliance, risk management and human rights-oriented development and deployment of AI.

**2. The debate on AI regulation**

While data processing regulation has been focused for decades on the law, including the interplay between data use and human rights, in recent years the debate on AI and the use of data-intensive systems has rapidly changed its trajectory, from law to ethics.[5] This is evident not only in the

---

*Norms and Domestic Change* (Cambridge University Press 1999) 234-278; Declan O'sullivan, 'The History of Human Rights across the Regions: Universalism vs Cultural Relativism' (1998) 2 The International Journal of Human Rights 22-48. However, from a policy and regulatory perspective, we believe that the human rights framework, including its nuances, can provide a more widely applicable common framework than other context-specific proposals on the regulation of the impact of AI. Furthermore, the proposed methodology includes in its planning section (see Section 5.1) the analysis of the human rights background, with a contextualisation based on local jurisprudence and laws, as well as the identification and engagement of potential stakeholders who can contribute to a more context-specific characterisation of the human rights framework.

[5] See Charles D Raab, 'Information Privacy, Impact Assessment, and the Place of Ethics' in this Review, 2020, 37, DOI: 10.1016/j.clsr.2020.105404, para 3.

literature,[6] but also in the political and institutional debate.[7] In this regard, an important turning point was the EDPS initiative on digital ethics[8] which led to the creation of the Ethics Advisory Group.[9]

As regards the debate on data ethics, it is interesting to consider its origins. We can identify three different and chronologically consecutive stages: the academic debate, institutional initiatives, and the proliferation of AI ethical codes.[10] These contributions to the debate are different and have given voice to different underlying interests.

The academic debate on the ethics of machines is part of the broader and older reflection on ethics and technology. It is rooted in known and framed theoretical models, mainly in the philosophical domain, and has a methodological maturity. In contrast, the institutional initiatives are more recent, have a non-academic nature and aim at moving the regulatory debate forward, including ethics in the sphere of data protection. The main reason for this emphasis on ethics in recent years has been the growing concern in society about the use of data and new data-intensive applications, such as Big Data[11] and, more recently, AI.

Although similar paths are known in other fields, the shift from the theoretical analysis to the political arena represents a major change. The political attention to these issues has necessarily reduced the level of analysis, ethics being seen as an issue to be flagged rather than developing a full-blown strategy for ethically-oriented solutions. In a nutshell, the message of regulatory bodies to the technology environment was this: law is no longer enough, you should also consider ethics.

This remarkable step forward in considering the challenges of new paradigms had the implicit limitation of a more general and basic ethical framework, compared to the academic debate. In some cases, only general references to the need to consider ethical issues has been added to AI strategy documents, leaving the task of further investigation to the recipients of these documents. At other times, as in the case of the EDPS, a more ambitious goal of providing ethical guidance was pursued.

---

[6] See e.g. Floridi Luciano and Taddeo Mariarosaria, 'What is data ethics?' (2016) 374(2083) Phil. Trans. R. Soc. A., DOI: 10.1098/rsta.2016.0360.

[7] In the context of the legal debate on computer law, at the beginning of this decade only few authors focused on ethical impact of IT, see e.g. David Wright, 'A framework for the ethical impact assessment of information technology' (2010) 13 Ethics Inf. Technol. 199–226. Although the reflection on ethics and technology is not new in itself, it has become deeper in the field of data use where new technology development in the information society has shown its impact on society. See also Peter-Paul Verbeek, *Moralizing Technology. Understanding and Designing the Morality of Things* (The University of Chicago Press 2011); Sarah Spiekermann, *Ethical IT Innovation: A Value-Based System Design Approach* (CRC Press 2016); J Bohn and others, 'Social, Economic, and Ethical Implications of Ambient Intelligence and Ubiquitous Computing' in Werner Weber, Jan M Rabaey and Emile Aarts (eds), *Ambient Intelligence* (Springer 2005), 5, 19-29.

[8] See EDPS, 'Opinion 4/2015. Towards a new digital ethics: Data, dignity and technology', 11 September 2015.

[9] See EDPS, 'Decision of 3 December 2015 establishing an external advisory group on the ethical dimensions of data protection ('the Ethics Advisory Group')', 2016/C 33/01 OJEU.

[10] See Marcello Ienca and Effy Vayena, 'AI Ethics Guidelines: European and Global Perspectives' in Council of Europe, *Towards regulation of AI systems. Global perspectives on the development of a legal framework on Artificial Intelligence systems based on the Council of Europe's standards on human rights, democracy and the rule of law* (Council of Europe, DGI (2020)16), 42-64.

[11] See also Council of Europe, Consultative Committee of the Convention 108 (T-PD), 'Guidelines on the protection of individuals with regard to the processing of personal data in a world of Big Data', Strasbourg, 23 January 2017 T-PD(2017)01.

Methodologically, the latter goal has often been pursued by delegating the definition of guidelines to committees of experts, including some forms of wider consultation. As in the tradition of expert committees, a key element of this process is the selection of experts.

These committees were not only composed of ethicists or legal scholars but had a different or broader composition defined by the appointing bodies.[12] Their heterogeneous nature made them more similar to multi-stakeholder groups.

Another important element of these groups advising policymakers concerns their internal procedures: the actual amount of time given to their members to deliberate, the internal distribution of assigned tasks (in larger groups this might involve several sub-committees with segmentation of the analysis and interaction between sub-groups), and the selection of the rapporteurs. These are all elements that have an influence in framing the discussion and its results.

All these considerations clearly show the differences between the initial academic debate on ethics and the same debate as framed in the context of institutional initiatives. Moreover, this difference concerns not only structure and procedures, but also outcomes. The documents produced by the experts appointed by policymakers are often minimalist in terms of theoretical framework and focus mainly on the policy message concerning the relevance of the ethical dimension.

The variety of the ethical approaches, the lack of clear indications on the frame of reference or the reasons for preferring a certain ethical framework make it difficult to understand the key choices on the proposed ethical guidelines.[13] Moreover, the local perspective of the authors of these documents, in line with the context-dependent nature of ethical values, undermines the ambition to provide global standards or, where certain values are claimed to have general relevance, may betray a risk of ethical colonialism.

**3. Framing the ethical and the human rights-based approaches**

From the outset, the debate on data ethics has been characterised by an improper overlap between ethics and law, in particular with regard to human rights. In this sense, it has been suggested that ethical challenges should be addressed by "fostering the development and applications of data science while ensuring the respect of human rights and of the values shaping open, pluralistic and tolerant information societies".[14] We can summarise this approach as 'ethics first': ethics plays a central role in technology regulation because it is the root of any regulatory approach, the pre-legal humus that is more important than ever where existing rules do not address or only partially address technological challenges.

Another argument in favour of the central role of ethics comes out of what we might call the 'ethics after' approach.[15] In the concrete application of human rights we necessarily have to balance competing interests. This balance test is not based on the rights themselves but on the underlying ethical values, meaning that the human rights framework is largely incomplete without ethics.

---

[12] This is the case, for example, of the Independent High-Level Expert Group on Artificial Intelligence set up by the European Commission, which brought together 52 experts, the majority (27) from industry and the rest from academia (15, including 3 with a legal background and 3 with an ethical background), civil society (6) and governmental or EU bodies (4). See also Access Now, 'Laying down the Law on AI: Ethics Done, Now the EU Must Focus on Human Rights' (8 April 2019) <https://www.accessnow.org/laying-down-the-law-on-ai-ethics-done-now-the-eu-must-focus-on-human-rights/> accessed 7 April 2021; Michael Veale, 'A Critical Take on the Policy Recommendations of the EU High-Level Expert Group on Artificial Intelligence' (2020) *European Journal of Risk Regulation,* 1-10. doi:10.1017/err.2019.65.
[13] See also Ienca and Vayena (fn 10).
[14] Floridi and Taddeo (fn 6).
[15] See Cansu Canca, 'AI & Global Governance: Human Rights and AI Ethics - Why Ethics Cannot Be Replaced by the UDHR' (2019) UN University Centre for Policy Research <https://cpr.unu.edu/ai-global-

Both these approaches are only partially correct. It is true that human rights have their roots in ethics. There is an extensive literature on the relationship between ethics and law, which over the years has been described by various authors as identification, separation, complementation, and interweavement.[16] Similarly, the influence of ethical values and more in general of societal issues in court decisions and balancing tests is known and has been investigated by various disciplines, including sociology, law & economics, and psychology.

Here the point is not to cut off the ethical roots, but to recognise that rights and freedoms flourish on the basis of the shape given them by law provisions and case law. There is no conflict between ethical values and human rights, but the latter represent a specific crystallisation of these values that are circumscribed and contextualised by legal provisions and judicial decisions.

This reflection may lead to a broader discussion of the role of ethics in the legal realm, but this study takes a more pragmatic and concrete approach by reframing the interplay between these two domains within the context of AI and focusing on the regulatory consequences of adopting an approach focused on ethics rather than human rights.

The main question should be formulated as follows: what are the consequences of framing the regulatory debate around ethical issues? Four different consequences can be identified: (1) uncertainty, (2) heterogeneity, (3) context dependence, and (4) risks of overestimation and of a 'transplant' of ethical values.

As far as uncertainty is concerned, this is due to the improper overlap between law and ethics in ethical guidelines.[17] While it is true that these two realms are intertwined in various ways, from a regulatory perspective the distinction between ethical imperatives and binding provisions is important. Taking a pragmatic approach, relying on a framework of general ethical values (such as beneficence, non-maleficence, etc.), on codes of conduct and ethical boards is not the same as adopting technological solutions on the basis of binding rules.

This difference is not only due to the different levels of enforcement, but also to the more fundamental problem of uncertainty about specific requirements. Stating that "while many legal obligations reflect ethical principles, adherence to ethical principles goes beyond formal compliance with existing laws"[18] is not enough to clarify the added value of the proposed ethical principles and their concrete additional regulatory impact.[19]

Given the different levels of binding nature and enforcement, shifting the focus from law to ethics and reformulating legal requirements as ethical duties open the door to de-regulation and self-

---

governance-human-rights-and-ai-ethics-why-ethics-cannot-be-replaced-by-the-udhr.html> accessed 30 April 2020.

[16] See Adela Cortina, 'Legislation, Law and Ethics' (2000) 3 Ethical Theory and Moral Practice 3-7.

[17] See Raab (fn 5) ("The products in the 'turn' to ethics often look more like 'data protection-plus' than a different kind of encounter with some of the age-old issues and concepts in the study and practice of ethics, and how to embed them in practice").

[18] Independent High-Level Expert Group on Artificial Intelligence set up by the European Commission (hereinafter AI HLEG), 'Ethics Guidelines for Trustworthy AI' (2019), 12.

[19] See for example the principle of respect for human autonomy, AI HLEG (fn 18), which is specified in self-determination, democratic process and human oversight. These are general categories that have an ethical origin but already have a concrete legal implementation – from personality rights to product safety – which can provide a better and more detailed framework for the elaboration of contextualised provisions concerning AI.

regulation. Rather than binding rules, business can therefore benefit from a more flexible framework based on corporate codes of ethics.[20]

This generates uncertainty in the regulatory framework. When ethical guidelines refer to human oversight, safety, privacy, data governance, transparency, diversity, non-discrimination, fairness, and accountability as key principles, they largely refer to legal principles that already have their contextualisation in specific provisions in different fields. The added value of a new generalisation of these legal principles and their concrete applications is unclear and potentially dangerous: product safety and data governance, for instance, should not be perceived as mere ethical duties, but companies need to be aware of their binding nature and related legal consequences.

Moreover, ethical principles are characterised by an inherent heterogeneity due to the different ethical positions taken by philosophers over the centuries. Virtue ethics, deontological or consequentialist approaches[21] can lead to different conclusions on ethical issues. AI developers or manufacturers might opt for different ethical paradigms (note that those mentioned are limited to the Western tradition only), making harmonised regulation difficult.

Similarly, the context-dependence of ethical values entails their variability depending on the social context or social groups considered, as well as the different ethical traditions.

By contrast, although the universal nature of human rights necessarily entails contextualised application through national laws, which partially create context dependency and can lead to a certain degree of heterogeneity, human rights seem to provide a more stable framework. The different charters, with their provisions, but also regional courts (such as the European Court of Human Rights), and a coherent legal doctrine based on international experience can all help to reduce this dependence on context.

This does not mean that human rights do not present contextual differences, but compared with ethical values, they are clearer, better defined, and stable. From a regulatory perspective, this facilitates a better harmonisation and reduces the risk of uncertainty.

A fourth important, and largely unaddressed issue in the current debate on AI and ethics concerns the methodological approach that we might call overestimation and 'transplant' of ethical values. In a context characterised by ethics guidelines that pop up "like woodland mushrooms in a wet Autumn",[22] a number of studies have attempted to identify the core values through a quantitative and text-based approach.[23]

The limitations of these studies are not restricted to the use of grey literature, results from search engines, and linguistic biases. The main limitation affecting this quantitative text-based approach is the lack of a policy perspective and contextual analysis.

Differing sources are considered on the same level, without taking into account the difference between the guidelines adopted by governmental bodies, independent authorities, private or public ad hoc committees, big companies, NGOs, academia, intergovernmental bodies etc. The mere frequency of occurrence does not reveal the different impacts of the distribution of these values among the different categories. For instance, the fact that some values are shared by several intergovernmental documents may have a greater policy impact than the same frequency in a cluster of NGOs or academic documents. When the focus is on values for future regulations, albeit based on ethics, the varying relevance of the sources in terms of political impact is important.

---

[20] See Ben Wagner, 'Ethics as an escape from regulation: From "ethics-washing" to ethics-shopping?' in Emre Bayamlıoğlu et al (eds), *Being Profiled* (Amsterdam University Press 2018) 84-89; Linnet Taylor and Lina Dencik, 'Constructing Commercial Data Ethics' (2020) Technology and Regulation 1-10, DOI: 10.26116/techreg.2020.001; Ienca and Vayena (fn 10).
[21] See Verbeek (fn 7), 30-33 and 61-63.
[22] See Raab (fn 5).
[23] See e.g. Anna Jobin, Marcello Ienca and Effy Vayena, 'The Global Landscape of AI Ethics Guidelines' (2019) 1 Nature Machine Intelligence 389–399. The authors identified ten key ethical values in a set of 84 policy documents with the following distribution: transparency 73/84; non-maleficence 60/84; responsibility 60/84; privacy 47/84; beneficence 41/84; freedom and autonomy 34/84; trust 28/84; sustainability 14/84; dignity 13/84, and solidarity 6/84.

Despite this limitation, seven values are present in most documents:[24] five of them are ethical values with a strong legal implementation (transparency, responsibility, privacy, freedom and autonomy) and only two come from the ethical discourse (non-maleficence and beneficence).

Another study[25] identified several guiding values and the top nine, with a frequency of 50% or more, are: privacy protection; fairness, non-discrimination and justice; accountability; transparency and openness; safety and cybersecurity; common good, sustainability and well-being; human oversight, control and auditing; solidarity, inclusion and social cohesion; explainability and interpretability. As in the previous study, the aggregating of these principles is necessarily influenced by the categories used by the authors to reduce the variety of principles. In this case, if we exclude values with a legal implementation, the key ethical values are limited to openness, the common good, well-being and solidarity.

If we take a qualitative approach, restricting the analysis to the document adopted by the main European organisations and to those documents with a general and non-sectoral perspective,[26] we can better identify the key values that are most popular among rule makers.

Considering the four core principles (respect for human autonomy, prevention of harm, fairness, and explicability) identified by the High-Level Expert Group on Artificial Intelligence,[27] respect for human autonomy and fairness are widely developed legal principles in the field of human rights and law in general, while explicability is more a technical requirement than a principle. Regarding the seven requirements identified by the HLGAI[28] on the basis of these principles, human agency and oversight are further specified as respect for fundamental rights, informed autonomous decisions, the right not to be subject to purely automated decisions, and adoption of oversight mechanisms. These are all requirements already present in the law in various forms, especially with regard to data processing. The same applies to the remaining requirements (technical robustness and safety, privacy and data governance; transparency; diversity, non-discrimination and fairness; accountability; and environmental well-being).

Looking at the entire set of values provided by the HLGAI, the only two elements – as framed in the document – that are only partially considered by the law are the principle of harm prevention – where "harms can be individual or collective, and can include intangible harm to social, cultural and political environments" – and the broad requirement of societal well-being, which generally requires a social impact assessment.

---

[24] See Jobin, Ienca and Vayena (fn 23).
[25] Thilo Hagendorff, 'The Ethics of AI Ethics: An Evaluation of Guidelines' (2020) 30 Minds and Machines 99, 102.
[26] See e.g. Council of Europe – CEPEJ, 'European Ethical Charter on the Use of Artificial Intelligence in Judicial Systems and their environment', Strasbourg, 3-4 December 2018.
[27] AI HLEG (fn 18).
[28] These requirements are: human agency and oversight; technical robustness and safety; privacy and data governance; transparency; diversity, non-discrimination and fairness; societal and environmental wellbeing; accountability.

Another important EU document identifies nine core ethical principles and democratic prerequisites.[29] Among them, four have a broader content that goes beyond the legal context (human dignity, autonomy, solidarity and sustainability). However, in the field of law and technology, human dignity and autonomy are two key values widely considered both in the human rights framework and in specific legal instruments.

Based on the results of these different analytical methodologies (quantitative, qualitative), we can identify two main groups of values that expand the legal framework. The first consists of broad principles derived from ethical and sociological theory (common good, well-being, solidarity). These principles can play a crucial role in addressing societal issues concerning the use of AI, but their broad nature might be a limitation if they are not properly investigated and contextualised. The most interesting group is the second one, which includes the principle of non-maleficence, the principle of beneficence,[30] and the related general claim of harm prevention. These are not new and undefined principles, especially in the field of applied ethics and research and medical ethics. In regard to these, we should therefore consider the potential risk of the 'transplant' of ethical values.[31]

It is not surprising that both the AI HLEG and the EU Commission,[32] when deciding to concretise the suggested ethical approach, moved from ethics to human rights. This is in line with the model suggested in the literature – and supported by NGOs active in the field of human rights – that considers human rights as the core of future AI regulation.[33] Compared with this doctrinal approach, the conclusion reached in Europe by the AI HLEG and the Commission still shows a partially improper overlap between these two areas.[34]

---

[29] European Commission - European Group on, Ethics in Science and New Technologies, 'Statement on Artificial Intelligence, Robotics and "Autonomous" Systems' (2018) <https://ec.europa.eu/research/ege/pdf/ege_ai_statement_2018.pdf> accessed 11 March 2018. The Ethical principles and democratic prerequisites identified are: human dignity; autonomy; responsibility; justice, equity, and solidarity; democracy; rule of law and accountability; security, safety, bodily and mental integrity; data protection and privacy; sustainability.

[30] Although in the final version of the AI HLEG (fn 18) these two principles are not explicitly listed as key values, they do underpin the whole approach of the AI HLEG, as demonstrated by the text of the draft version of the guidelines used for the public consultation; see AI HLEG, 'Draft Ethics Guidelines for Trustworthy AI' (2018) <https://ec.europa.eu/digital-single-market/en/news/draft-ethics-guidelines-trustworthy-ai> accessed 18 December 2018 ("Ensure that AI is human-centric: AI should be developed, deployed and used with an "ethical purpose", grounded in, and reflective of, fundamental rights, societal values and the ethical principles of Beneficence (do good), Non-Maleficence (do no harm), Autonomy of humans, Justice, and Explicability. This is crucial to work towards Trustworthy AI.").

[31] See also Zachary M Schrag, *Ethical Imperialism. Institutional Review Boards and the Social Sciences 1965-2009* (Johns Hopkins University Press 2017).

[32] See AI HLEG (fn 18); European Commission, 'White paper on Artificial Intelligence - A European approach to excellence and trust' COM(2020) 65 final, Brussels, 19 February 2020; European Parliament, 'Framework of ethical aspects of artificial intelligence, robotics and related Technologies European Parliament resolution of 20 October 2020 with recommendations to the Commission on a framework of ethical aspects of artificial intelligence, robotics and related technologies (2020/2012(INL))' P9_TA-PROV(2020)0275.

[33] See Alessandro Mantelero, 'Regulating AI within the Human Rights Framework: A Roadmapping Methodology' in Philip Czech et al. (eds) *European Yearbook on Human Rights* (Intersentia, 2020) 477-502; Karen Yeung, Andrew Howes and Ganna Pogrebna, 'AI Governance by Human Rights–Centered Design, Deliberation, and Oversight' in Markus D. Dubber, Frank Pasquale, and Sunit Das (eds), *The Oxford Handbook of Ethics of AI* (Oxford University Press 2020), 77-106; Lorna McGregor, Daragh Murray and Vivian Ng, 'International Human Rights Law as a Framework for Algorithmic Accountability' (2019) 68 International and Comparative Law Quarterly 309-343; Access Now, 'Human Rights in The Age of Artificial Intelligence' (2018) <https://www.accessnow.org/cms/assets/uploads/2018/11/AI-and-Human-Rights.pdf> accessed 23 November 2020; Paul Nemitz, 'Constitutional Democracy and Technology in the Age of Artificial Intelligence' (2018) 376 Philosophical Transactions of the Royal Society A: Mathematical, Physical and Engineering Sciences, DOI: /10.1098/rsta.2018.0089.

The notion of solving the problems of the variability of ethical values and their contextual nature by linking them to human rights seems inappropriate. Ethics can play an important role in AI regulation, not as a backdrop to a human rights-based approach, creating confusion about the existing definition of these rights, but as a complementary element. Ethics can cover those issues that are not addressed, or not fully addressed, by the human rights framework, and revolve around a discretionary evaluation based on the socio-ethical values of a given community with respect to the various human rights-oriented options to be chosen.[35]

For example, a heavily data-driven community (see Section 6) is not necessarily contrary to human rights, since data can be processed in compliance with the law. But it does raise questions as to how we wish technology to be used in society, to what extent we want a pervasive digitalisation of everything in the name of a promised greater efficiency. These are ethical and social quandaries that need to be addressed and go beyond the human rights framework.

Based on these considerations on the interplay between law and ethics in regulating AI, for the purposes of this contribution, we can conclude that a focus on the human rights framework is necessary and crucial for an effective development of a human-centric AI. In this regard, an *ex ante* Human Rights Impact Assessment (HRIA)[36] is a necessary requirement in the development and deployment of AI solutions to prevent any prejudice to human rights and fundamental freedoms, as well as to promote a human rights-oriented AI.[37]

At the same time, a positive outcome of such an assessment does not exclude the presence of ethical issues related to the proposed solution, which should be further investigated.[38] The HRIA thus makes the role of human rights in technology development evident and, in this way, helps to avoid improper overlap with ethical issues.

**4. Defining an operational approach to human rights assessment in AI**

---

[34] This uncertain focus is still present in the recent European Parliament (fn 32), where the core requirements are based on fundamental rights, but the framework refers to ethics, including on risk management.
[35] See Raab (fn 5); Alessandro Mantelero, 'AI and Big Data: A blueprint for a human rights, social and ethical impact assessment', in this Review, 2018, 34(4), 754-772.
[36] HRIA differs from other assessment methodologies used in the field of technology, such as technology assessment (TA) or impact assessment (IA). TA and IA have different focuses: TA is centred on technology development (e.g. road mapping) while IA on policy planing. See also Armin Grunwald, 'Technology Assessment: Concepts and Methods', in Anthonie Meijers, *Philosophy of Technology and Engineering Sciences. Handbook of the Philosophy of Science*, vol. 9 (Elsevier, 2009), 1103-1146; Thien A. Tran and Tugrul Daim, 'A taxonomic review of methods and tools applied in technology assessment' (2008) 75(9) Technol. Forecast. Soc. Change 1396-1405; World Bank and Nordic Trust Fund, 'Human Rights Impact Assessments: A Review of the Literature, Differences with other forms of Assessments and Relevance for Development' (Washington, 2013).
[37] See Council of Europe – Ad hoc Committee on Artificial Intelligence, 'Feasibility Study', Strasbourg, 17 December 2020, <www.coe.int/cahai> accessed 23 January 2021, 44, 50; European Union Agency for Fundamental Rights (FRA), 'Getting the Future Right – Artificial Intelligence and Fundamental Rights' (2020) <https://fra.europa.eu/en/publication/2020/artificial-intelligence-and-fundamental-rights> accessed 14 December 2020, 87-98.
[38] For this reason, models like the HRESIA (Mantelero, fn 33) may be better suited to grasp the holistic definition of the relationship between humans and machines.

In considering the impact of AI on human rights, the dominant approach in many documents is mainly centred on listing the rights and freedoms potentially impacted[39] rather than operationalising this potential impact and proposing assessment models.

Moreover, case-specific assessment is more effective in terms of risk prevention and mitigation than using risk presumptions based on an abstract classification of "high-risk sectors and high-risk uses or purposes",[40] where sectors, uses and purposes are very broad categories which include different kind of applications – some of them continuously evolving – with a variety of potential impacts on rights and freedoms that cannot be clustered ex ante on the basis of risk thresholds, but require a case-by-case impact assessment.

Similarly, the adoption of a centralised technology assessment carried out by national ad hoc supervisory authorities[41] can provide useful guidelines for technology development and can be used to fix red lines[42] but must necessarily be complemented by a case-specific assessment of the impact of each application developed.

For these reasons, a case specific impact assessment remains the main tool to ensure accountability and the safeguarding of individual and collective rights and freedoms. In this regard, a solution to the problem could easily be drawn from the human rights impact assessment models already adopted in several fields.

However, these models are usually designed for different contexts than those of AI applications.[43] The latter are not necessarily large-scale projects involving entire regions with multiple social impacts. Although there are important data-intensive projects in the field of smart cities, regional services (e.g. smart mobility) or global services (e.g. online content moderation provided by big players in social media), the AI operating context for the coming years will be more fragmented and distributed in nature, given the business environment in many countries, often dominated by SMEs, and the variety of communities interested in setting-up AI-based projects. The growing number of data scientists and the decreasing cost of hardware and software solutions, as well as their delivery as a service, will facilitate this scenario characterised by many projects with a limited scale, but involving thousands of people in data-intensive experiments.

---

[39] See Filippo Raso, Hannah Hilligoss, Vivek Krishnamurthy, Christopher Bavitz, Levin Kim, 'Artificial Intelligence & Human Rights Opportunities & Risks' (2018) <https://cyber.harvard.edu/sites/default/files/2018-09/2018-09_AIHumanRightsSmall.pdf?subscribe=Download+the+Report> accessed 28 September 2018; AI HLEG (fn 18); Council of Europe, Committee of Ministers, Recommendation CM/Rec(2020)1 on the human rights impacts of algorithmic systems; Council of Europe, 'Algorithms and Human Rights. Study on the Human Rights Dimensions of Automated Data Processing Techniques and Possible Regulatory Implications' (Council of Europe, 2018).

[40] See European Parliament (fn 32). The Annex (Exhaustive and cumulative list of high-risk sectors and of high-risk uses or purposes that entail a risk of breach of fundamental rights and safety rules) considers transport as a high-risk sector for AI application, but there are several AI applications in this field based on non-personal data (e.g. security systems for trains) and related to infrastructure management with no relevant impact on rights and freedoms. Similarly, energy production and distribution is considered as high-risk use/purpose, but several automated energy sharing and switching solutions used in this field have no relevant impacts on rights and freedoms. In addition, this proposal and its Annex combine the impact on fundamental rights impact and safety risks, which are different types of potential risks, based on different criteria and assessment methodologies, without clarifying which sector, use or purpose is relevant in terms of potential adverse impact on fundamental rights and where the impact is limited to safety only. Furthermore, a harm-based approach is taken ("cause injury or harm"), rather than a rights-based approach focusing on potential prejudice to fundamental rights and freedoms, centered on risk prevention and accountability, as in the human rights impact assessment and in the GDPR.

[41] See European Parliament (fn 32), Article 14.2 ("the risk assessment of artificial intelligence, robotics and related technologies, including software, algorithms and data used or produced by such technologies, shall be carried out, in accordance with the objective criteria provided for in paragraph 1 of this Article and in the exhaustive and cumulative list set out in the Annex to this Regulation, by the national supervisory authorities referred to in Article 18 under the coordination of the Commission and/or any other relevant institutions, bodies, offices and agencies of the Union that may be designated for this purpose in the context of their cooperation").

[42] See, on the debate on the adoption of specific red lines regarding the use of AI in the field of facial recognition, European Digital Rights (EDRi), 'Civil Society Calls for AI Red Lines in the European Union's Artificial Intelligence Proposal' (2021) <https://edri.org/our-work/civil-society-call-for-ai-red-lines-in-the-european-unions-artificial-intelligence-proposal/> accessed 15 March 2021.

[43] See below fn 153 and 245.

For such projects, the traditional HRIA models are too articulated and oversized, which is why it is important to provide a more tailored model of impact assessment, at the same time avoiding mere theoretical abstractions based on generic decontextualised notions of human rights. To address this challenge, we have chosen to build the proposed model on the experience of data protection authorities (hereafter DPAs) in Europe, taken as a case study to identify the impacted areas of data intensive systems in relation to human rights.

DPAs, more than any other supervisory or judicial body, have in the last few decades addressed crucial issues concerning the use of data-intensive and data-invasive systems. Moreover, the jurisprudence of these authorities, both at national and EU level, has traditionally been inspired by attention to respect for fundamental rights – also given the strict relationship existing in the European context between data protection and personality rights[44] – as confirmed by the nature of fundamental right recognised to data protection and the European Court of Human Rights jurisprudence on the protection of private life.[45]

Against this background, the following sub-sections will investigate the jurisprudence of these authorities to figure out how data-intensive systems potentially affect human rights.

Before passing to the empirical analysis providing solid evidence for a human rights assessment, it is worth briefly considering the role played by impact assessment tools with respect to the precautionary principle as an alternative way of dealing with the consequences of AI.

As in the case of potential technology-related risks, there are two different legal approaches to the challenges of AI: the precautionary approach and the risk assessment. These approaches are alternative, but not incompatible. Indeed, complex technologies with a plurality of different impacts might be better addressed though a mix of these two remedies.[46]

As risk theory states, their alternative nature is related to the notion of uncertainty.[47] Where a new application of technology might produce potential serious risks for individuals and society, which cannot be accurately calculated or quantified in advance, a precautionary approach should be taken.[48] In this case, the uncertainty associated with applications of a given technology makes it impossible to conduct a concrete risk assessment, which requires specific knowledge of the extent of the negative consequences, albeit in specific classes of risks.[49]

---

[44] See Gert Brüggemeier, 'Protection of personality rights in the law of delict/torts in Europe: mapping out paradigms' in Gert Brüggemeier, Aurelia Colombi Ciacchi and Patrick O'Callaghan, *Personality Rights in European Tort Law* (Cambridge University Press 2010), 5-37; Stig Stromhölm, *Right of Privacy and Rights of Personality. A comparative Survey* (Norstedt & Soners 1967) 28–31.

[45] See European Union Agency for Fundamental Rights and Council of Europe, *Handbook on European Data Protection Law* (2018) <http://fra.europa.eu/en/publication/2018/handbook-european-data-protection-law> accessed 25 May 2018.

[46] See also Council of Europe, Consultative Committee of the Convention 108 (T-PD) (fn 11), Section IV, paras 1 and 2, where the precautionary approach is coordinated with an impact assessment that also includes ethical and social issues.

[47] On the distinction between the precautionary approach and the precautionary principle, see Jacqueline Peel, 'Precaution - A Matter of Principle, Approach or Process?' (2004) 5(2) Melb. J. Int. Law 483 <http://www.austlii.edu.au/au/journals/MelbJlIntLaw/2004/19.html> accessed 4 February 2017 ("One way of conceptualising what might be meant by precaution as an approach […] is to say that it authorises or permits regulators to take precautionary measures in certain circumstances, without dictating a particular response in all cases. Rather than a principle creating an obligation to act to address potential harm whenever scientific uncertainty arises, an approach could give regulators greater flexibility to respond").

Where the potential consequences of AI cannot be fully envisaged, as in the case of the ongoing debate on facial recognitions and its applications, a proper impact assessment is impossible, but the potentially high impact on society justifies specific precautionary measures (e.g., a ban or restriction on the use of AI-based facial recognition technologies).[50] This does not mean limiting innovation, but investigating more closely its potentially adverse consequences and guiding the innovation process and research, including the mitigation measures (e.g. containment strategies, licensing, standards, labelling, liability rules, and compensation schemes).

On the other hand, where the level of uncertainty is not so high, the risk-assessment process is a valuable tool in tackling the risks stemming from technology applications. According to the general theory on the risk-based approach, the process consists of four separate stages: 1) identification of risks, 2) analysis of the potential impact of these risks, 3) selection and adoption of the measures to prevent or mitigate the risks, 4) periodic review of the effectiveness of these measures.[51] Furthermore, to enable subsequent monitoring of the effective level of compliance, duty bearers should document both the risk assessment and the measures adopted.

Since neither the precautionary principle nor the risk assessment are an empty list but rather focus on specific rights and freedoms to be safeguarded, they can be seen as two tools for developing a human rights-centred technology. While the uncertainty of some technology solutions will lead to the application of the precautionary principle, a better awareness and management of related risk will enable a proper assessment.

---

[48] Only few contributions in law literature take into account the application of the precautionary approach in the field of data protection, see Luiz Costa, 'Privacy and the precautionary principle' in this Review (2012), 28(1), 14–24 and Maria Eduarda Gonçalves, 'The EU data protection reform and the challenges of big data: remaining uncertainties and ways forward' (2017) 26(2) Inform. Comm. Tech. Law 90-115. See also Wolter Pieters. 'Security and Privacy in the Clouds: A Bird's Eye View' in Serge Gutwirth, Yves Poullet, Paul de Hert, Ronald Leenes, (eds.), *Computers, Privacy and Data Protection: an Element of Choice* (Springer 2011) 455 ("generalised to information technology, it can serve as a trigger for government to at last consider the social implications of IT developments. Whereas the traditional precautionary principle targets environmental sustainability, information precaution would target social sustainability"). On the precautionary approach in data protection, see also Arvind Narayanan, Huey Joanna and Edward W. Felten, 'A Precautionary Approach to Big Data Privacy' in Serge Gutwirth, Ronald Leenes, e Paul De Hert (eds) *Data Protection on the Move* (Springer 2016) 357-385; Charles Raab and David Wright, 'Surveillance: Extending the Limits of Privacy Impact Assessment' in David Wright and Paul De Hert (eds), *Privacy Impact Assessment* (Springer 2012) 364; Orla Lynskey, *The Foundations of EU Data Protection Law* (Oxford University Press 2015) 83; Charles Raab, 'The future of privacy protection' (2004) Cyber Trust & Crime Prevention Project 15 <https://www.piawatch.eu/node/86> accessed 28 April 2017.
[49] See also Jale Tosun, 'How the EU Handles Uncertain Risks: Understanding the Role of the Precautionary Principle' (2013) 20 (10) JEPP, 1517-1528; Terje Aven, 'On Different Types of Uncertainties in the Context of the Precautionary Principle' (2011) 31(10) Risk Analysis 1515–1525; Andy Stirling and David Gee, 'Science, precaution, and practice' (2002) 117(6) Public Health Reports 521–533.
[50] See e.g. European Parliament - Committee on Civil Liberties, Justice and Home Affairs, 'Opinion of the Committee on Civil Liberties, Justice and Home Affairs for the Committee on Legal Affairs on artificial intelligence: questions of interpretation and application of international law in so far as the EU is affected in the areas of civil and military uses and of state authority outside the scope of criminal justice' 2020/2013(INI) (2020), paras 14, 15 and 20. See also Council of Europe, Consultative Committee of the Convention 108 (T-PD), 'Guidelines on Facial Recognition', 28 January 2021, T-PD(2020)03rev4, para 1.1.
[51] See also Raija Koivisto and David Douglas, 'Principles and Approaches in Ethics Assessment. Ethics and Risk. Annex 1.h Ethical Assessment of Research and Innovation: A Comparative Analysis of Practices and Institutions in the EU and selected other countries. Project Stakeholders Acting Together on the Ethical Impact Assessment of Research and Innovation – SATORI. Deliverable 1.1' (2015) < http://satoriproject.eu/work_packages/comparative-analysis-of-ethics-assessment-practices/> accessed 15 February 2017.

However, the relationship between risk assessment and the precautionary principle is rather complicated and cannot be reduced to a strict alterative. Indeed, when a precautionary approach suggests that a technology should not be used in a certain social context, this does not necessary entail halting its development. On the contrary, where there is no incompatibility with human rights (e.g. mass destruction harms) the technology can be developed further to reach a sufficient level of maturity that shows awareness of the related risks and the effective solutions.

This means that, in these cases, human rights can play an additional role in guiding development such that, once it reaches a level of awareness of the potential consequences that exclude uncertainty, will be subject to risk assessment.

Under this reasoning, two different scenarios are possible. One in which the precautionary principle becomes an outright ban on a specific use of technology and the other in which it restricts the adoption of certain technologies but not their further development. In the latter case, a precautionary approach and a risk assessment are two different phases of the same approach rather than an alternative response.

*4.1 A methodological approach for an evidence-based model*

Having defined the importance of a human rights-oriented approach in AI data processing, there remains the methodological question of how to define the assessment benchmark.

Three different approaches are possible: (i) a top-down theoretical approach; (ii) an inferential approach, and (iii) a bottom-up empirical approach. The first was used in the analysis conducted by Raso et al.[52], in which various potentially affected rights are analysed on the basis of abstract scenarios grouped by sector-specific applications (risk assessments in criminal justice, credit scoring, healthcare diagnostics, online content moderation, recruitment and hiring systems, essay scoring in education).

The second approach was adopted by Fjeld et al.[53], inferring values from existing ethics and right-based documents on AI regulation. This approach is close to the empirical approach, but is dominated by a quantitative dimension, focusing on the frequency of certain principles, and overlooking the heterogeneity of the documents. As the documents are often declarations by governmental and non-governmental bodies, they have the nature of guidelines and directives rather than concrete descriptions of the existing state-of-the-art: they are more focused on To-Be rather than on As-Is. This means that the prevalence of certain principles and values does not necessarily demonstrate a concrete and effective implementation of them.

The third approach, used in this work, adopts an evidence-based methodology grounded on empirical analysis of cases decided by DPAs and guidelines provided by these authorities . More specifically, the idea is to move from the reasoning adopted by decision-makers in scrutinising data-centred applications and use this experience to better understand which rights and freedoms are relevant in practice.

---

[52] See Raso et al. (fn 39).
[53] Jessica Fjeld et al., 'Principled Artificial Intelligence: Mapping Consensus in Ethical and Rights-Based Approaches to Principles for AI' Berkman Klein Center for Internet & Society, 2020 <https://dash.harvard.edu/handle/1/42160420> accessed 15 January 2020.

The advantage and complementarity nature of this approach is its consistency with the already existing practices in the field of data protection, where DPAs are the supervisory bodies that are addressing and will further address the largest number of issues related to AI context. Using an empirical legal research methodology, this analysis focuses on what already exists in concrete practice and can thus be extended to cases concerning AI. In this sense it differs from the first approach, in its bottom-up nature, and from the second, as here the link between the As-Is and To-Be is stronger.

A model based on the empirical evidence can be better understood and used both by supervisory authorities and operators already accustomed to existing DPAs' jurisprudence and practice. Moreover, from a regulatory point of view, this approach is consistent with the recent worldwide growth of evidence-based policies.

From a more specific data protection standpoint, the European proposals for a future AI regulation focus on risk assessment without providing concrete models for this assessment. In this regard, both the main European legally binding instruments, Convention 108+[54] and GDPR,[55] refer to "rights and fundamental freedoms" (Convention 108+) and to "rights and freedoms of natural persons" (GDPR), without specifically identifying them. Similarly, the DPIA templates adopted by DPAs do not provide enough detail on this point.

An evidence-based model has the advantage of not resting on its authors' abstract vision, which may be coloured by their individual theoretical or cultural standpoint but based on decision-makers' concrete practices. While they too are necessarily affected by cultural influences, these decision-makers are the ones who will address the new cases concerning AI. Adopting this culture-specific perspective does not undermine the expected outcome but puts it in context.

At the same time, the evidence-based analysis adopted here maintains a general approach to the issues it deals with, without a case-specific focus – such as in other empirical studies concerning AI[56] – that cannot be generalised.

Regarding the potential limitations affecting this approach, it is true that almost all the cases decided by supervisory authorities do not directly concern AI, given the time lag between technology development and litigation, but the reasoning adopted in current data intensive cases can be considered as a useful proxy for the application of human rights in data-intensive systems based on AI.[57]

In addition, compared to the decision of international courts, and the European Court of Human Rights,[58] DPAs have implemented a broader set of actions, not circumscribed to case decisions, including guidelines and other documents that contribute to defining best practices in data processing and are of interest for the contextualisation of human rights and freedoms in the AI context.

---

[54] See Convention 108+, Articles 6 and 10.
[55] GDPR, Article 35.
[56] See e.g. Kate Crawford and Vladan Joler, 'Anatomy of an AI System: The Amazon Echo As An Anatomical Map of Human Labor, Data and Planetary Resources' (AI Now Institute and Share Lab, 7 September 2018) <http://www.anatomyof.ai> 20 October 2018.
[57] It is also worth noting that in the DPAs' decisions, a direct focus on human rights is less prominent and explicit in the motivations, as detailed below in Section 4.2.
[58] See e.g. the decision of the Court of Justice of the European Union on the so-called right to be forgotten and the further need to define specific policies for its implementation. See Court of Justice of the European Union, 13 May 2014, Case C-131/12, Google Spain SL and Google Inc. v. Agencia Española de Protección de Datos (AEPD) and Mario Costeja González, OJ C 212, 07.07.2014, 4. Similarly, the decisions of the European Court of Human Rights (hereinafter ECtHR), albeit based on concrete cases, focus on rights and freedoms violations, but do not develop concrete solutions for data processing.

Given the enormous number of decisions made by these bodies, selection favoured cases where data use might entail an impact on human rights. The geographical area of investigation was limited to Europe, focussing on six countries with a longer experience in regulating data processing: Belgium, France, Germany, Italy, United Kingdom, and Spain.

Given the influence that social contexts may have on human rights, the selection took into account both authorities belonging to systems with a similar socio-cultural environment (e.g. Italy and Spain), and authorities belonging to systems with a distinct legal and social culture (UK). Since this necessarily entailed the exclusion of a significant area within the EU, the research also considered the opinions and documents adopted by the Article 29 Data Protection Working Party and the European Data Protection Board to have a more comprehensive overview.

The decision not to circumscribe the research to the documents adopted by national DPAs was also influenced by the relationship between data protection claims and technology development. Indeed, the potential harms resulting from the use of innovative technologies and data-intensive systems might not yet be known to data subjects, but the prejudices may be well perceived and discussed in the context of the activities carried out by supranational bodies. This also explains why the decisions of national DPAs examined rarely refer to new technological solutions.

More than 700 documents were analysed,[59] selected on the basis of their relevance to human rights and fundamental freedoms. The concepts used to extract the most relevant documents from the databases of the decisions adopted by DPAs took into account, *inter alia*, the nature of the devices used for data collection (e.g. video-surveillance systems, geolocation tools, IoT systems and personal devices), the places where data is collected (public or private spaces, the workplace, etc.) and the nature of the data (e.g. biometric data).

The number and nature of documents examined differ on national basis. As these documents cover a wide period (1994-2020), most of the materials examined are based on Directive 95/46/EC and a more limited number of decisions refer to the GDPR, given its relatively recent entry into force and the inevitable time lag between the first implementation of a new law and decided cases.

Regarding the materials based on Directive 95/46/EC, it is worth noticing the heterogeneity of the documents as a result of the different powers exercised by the national authorities before the GDPR came into force[60], the different nature of their acts and their policy approaches.[61] This diversity has been mitigated by the advent of the GDPR, which provided for uniform regulation of the DPAs' powers, as reflected in the subset of decisions referring to the most recent cases.

---

[59] The analysis is based on the documentation made available on the official websites of the DPAs and the EDPB, see <https://www.garanteprivacy.it/> (Italian DPA); <https://www.cnil.fr> (French DPA); <https://www.autoriteprotectiondonnees.be/> (Belgian DPA); <https://www.bfdi.bund.de/DE/> (Federal German DPA); <https://ico.org.uk> (UK DPA); <https://edpb.europa.eu/edpb_en> (EDPB). The documents adopted by the Article 29-Data Protection Working Party are available at <https://ec.europa.eu/justice/article-29/documentation/index_en.htm>.

[60] See, Baker & Mc Kenzie, 'Global Data Protection Enforcement Report - Enforcement by regulators: penalties, powers and risks' (2016) <https://www.bakermckenzie.com/-/media/files/insight/events/2018/04/gdrp-enforcement.pdf?la=en>, accessed 10 December 2017.

[61] The documents from the Italian, Spanish and French authorities are mainly decisions on specific complaints, whereas the UK DPA documents tend to be guidelines, recommendations and information provided to various industries. The documents from Belgium are mainly recommendations. As regards Germany, the statements of the Federal DPA and the minutes of the meetings between the federal and the länder DPAs were taken into consideration. These meetings (Konferenz der unabhängigen Datenschutzbehörden des Bundes und der Länder) adopt agreed resolutions which outlining the attitude of federal and länder privacy authorities with regard to technical, economic and legal issues concerning data processing.

From a methodological perspective, although the materials examined can be divided into these two subsets (before and after the GDPR), this does not affect the overall analysis for several reasons: (i) Directive 95/46/EC and the GDPR are grounded on the same core principles; (ii) human rights are often relevant in data processing activities and in this sense data protection is considered as an enabling right with a view to human rights protection; (iii) although the GDPR introduced a risk assessment approach focused on the rights and freedoms of natural persons, the Regulation does not provide specific rules on rights and freedoms other than data protection and even the data protection impact assessment enshrined in Article 35 needs specific implementation.

For these reasons, we examined these two clusters of data in a unified way, without distinguishing between the pre- and post-GDPR periods. Moreover, for the EU context, this analysis can provide a contribution in terms of developing concrete solutions to carry out the impact assessment required by the GDPR.

As for document selection, the collected materials were analysed in detail to identify the most significant cases and discard those concerning the same issues or adopting a similar argumentative logic. At the end of this screening phase, 350 documents were taken into consideration for the purposes of this study (broken down as follows: Italy100, Spain 35, France 60, Belgium 40, Germany 20, United Kingdom 45, Article 29 Data Protection Working Party 50).[62]

*4.2 Human rights and data use in the DPAs' jurisprudence*

Despite the authorities considered belonging to different legal and cultural traditions, the analysis of the documents did find common ground between them in their approach to human rights and freedoms.

It worth noting that, whereas the importance of these interests is clearly stated, in several cases the analysis of their relevance is not properly developed and in others emerges only indirectly in the DPAs' observations. In fact, DPAs often prefer to refer to principles such as proportionality, necessity or transparency set forth in the data protection regulations to safeguard interests other than privacy and data protection, without a further elaboration.

---

[62] It is worth noting that this uneven distribution of cases among the countries examined is the result of a content-based approach and not of an intentional deeper investigation in one country rather than another. Several factors may have produced this (unintentional) geographical distribution due to their impact on publicly available decisions. For example, prior to the GDPR (the examined period is 1994-2020) some DPAs had no sanctioning power (UK, Belgium), some had concurrent competence with regional authorities (Germany), some DPAs (Belgium) had a model more favourable to ADRs, and some had less performing internal search engines. With regard to the latter, we conducted an initial selection based on keywords and, as there is no single search engine, but each authority has its own, the low performance of some of these search services may have provided fewer results for some DPAs. In addition, the issues investigated with respect to human rights are more present with respect to some topics (e.g. video surveillance) than others, with a potential different distribution among DPAs according to the concrete demands addressed. The fact that some DPAs have mainly issued recommendations also contributes to reducing the number of documents available for some DPAs, compared to others that have a large number of decided cases. Finally, even when DPAs have adopted many decisions, there are large clusters of cases concerning aspects (e.g. access rights or lack of informed consent) that have little or no relevance from a human rights perspective.

However, we identified a special attention to the possible risks for individual rights and freedoms, with specific reference to human rights principles and to several human rights and freedoms. The following sub-sections will analyse this evidence.

*4.2.1 Respect for human dignity*

A first core value underlying the DPA's decisions is human dignity, recognised as crucial in many legal systems and widely protected in European[63] and international frameworks.[64] Despite the difficulties in determining the precise meaning of this concept,[65] it is a key notion in human rights law and a guiding principle that underpins and grounds all other principles in human rights,[66] even in the context of data processing.[67]

We also found this broad notion of human dignity in the decisions of the DPAs, according to which human dignity encompasses various aspects of the individual sphere and is an important factor in many different contexts.

For instance, the DPAs recognise that negative outcomes for individual dignity may result from continuous and invasive monitoring, such as video surveillance or other monitoring technologies[68] or data-intensive systems collecting mobility data and driving behavioural information (e.g. GPS; Wi-Fi tracking devices; RFID technologies; Intelligent Transport Systems and "event data recorder" devices).[69] They regard these practices as potentially oppressive or demeaning, if associated risks are not mitigated.

---

[63] See Article 1, EU Charter of Fundamental Rights. See also, e.g., Court of Justice of the European Union, 12 November 2019, Case C-233/18, Zubair Haqbin v. Federaal agentschap voor de opvang van asielzoekers. Although the European Convention on Human Rights does not explicitly refer to the notion of human dignity, there is no doubt that, implicitly, this document affirms the value of respect for human dignity. A confirmation of this can be found in the jurisprudence of the European Court of Human Rights, where it is clear that human dignity is implicated in the Convention's protective regime (see e.g. ECtHR, 22 November 1995, S.W. v. the United Kingdom, App. No. 20166/92, para 44). See, among others, Roger Brownsword, 'Human dignity from a legal perspective', in Marcus Düwell, Jens Braarvig, Roger Brownsword and Dietmar Mieth (eds), *The Cambridge Handbook of Human Dignity. Interdisciplinary Perspectives* (Cambridge University Press 2014), 1.

[64] See UDHR, Preamble, which refers to dignity as an inherent value of each human being, simply as an innate consequence of human existence. In particular, dignity is seen as the core value that underpins human rights to which three basic values refer: liberty, equality and solidarity. See Bas De Gaay Fortman, 'Equal dignity in international human rights, in Düwell et al. (fn 63), 356. See also, among others, Jan Mårtenson, 'The Preamble of the Universal Declaration of Human Rights and the UN Human Rights Programme' in Eide Asbjørn et al. (eds), *The Universal Declaration of Human Rights: A Commentary* (Scandinavian University Press, 1992), 17.

[65] See, among others, Catherine Dupré, 'Art. 1 – Human dignity' in Steve Peers, Tamara Hervey, Jeff Kenner, and Angela Ward (eds), *The EU Charter of Fundamental Rights. A Commentary* (Hart Publishing 2014), 18. See also Roald Dworkin, *Taking Rights Seriously* (Harvard University Press 1978), 198-199.

[66] See Govert Den Hartogh, 'Is human dignity the ground of human rights?' in Düwell et al. (fn 63), 200; De Gaay Fortman (fn 64) 356; Dupré (fn 65) 24. See also *Explanation on Article 1 — Human Dignity (Explanations Relating to the Charter of Fundamental Rights)*, in Official Journal of the European Union C 303/17 - 14.12.2007 <https://eur-lex.europa.eu/legal-content/EN/TXT/PDF/?uri=OJ:C:2007:303:FULL&from=EN> accessed 2 September 2019 ("The dignity of the human person is not only a fundamental right in itself but constitutes the real basis of fundamental rights").

[67] See Council of Europe, Convention 108+, Preamble.

Human dignity also plays a role in DPAs' decisions on monitoring activities of a private or intimate nature which might create discomfort for individuals (e.g. monitoring of employees' electronic communications or Internet use[70]). According to the DPAs, human dignity is also particularly relevant when video-surveillance or other monitoring tools are used in environments characterised by a high privacy expectation (e.g. restrooms or changing rooms).[71]

Regarding the nature of the information used, DPAs see sensitive data as more closely linked to human dignity. This is evidenced by cases of invasive information requests by employers (e.g. health conditions, religious beliefs, criminal records, and drug and alcohol use),[72] biometric data collection,[73] and the use of wearable and IoT devices to gather sensitive data (e.g. health data) or profiling information.[74]

---

[68] With reference to video surveillance of workers, see Information Commissioner's Office (hereinafter ICO), *The employment practices code*, 2011, Part. 3; Garante per la protezione dei dati personali (hereinafter GPDP), 4 April 2013, doc. web n. 2439178; GPDP, 30 October 2013, doc. web n. 2851973; Commission de la protection de la vie privée (hereinafter CPVP), avis, n. 8/2006, 12 April 2006; Commission Nationale de l'Informatique et des Libertés (hereinafter CNIL) n. 2014-307, 17 July 2014; Der Bundesbeauftragte für den Datenschutz und die Informationsfreiheit (hereinafter BFDI), 'Videoüberwachung am Arbeitsplatz'. In relation to invasive monitoring activities see also GPDP, 25 January 2018, doc. web n. 7810766 (monitoring of patients, within a healthcare facility, through the use of wearable devices); Konferenz der unabhängigen Datenschutzbehörden des Bundes und der Länder (hereinafter DSK), '„Videoüberwachungsverbesserungsgesetz" zurückziehen!', 9 November 2016.

[69] See GPDP, 8 September 2016, doc. web n. 5497522; GPDP, 7 November 2013, n. 499, doc. web n. 2911484; CPVP, avis n. 12/2005, 7 September 2005; CPVP, recommandation n. 01/2010, 17 March 2010; CNIL n. 2010-096, 8 April 2010; Agencia Española de Protección de Datos (hereinafter AEPD), resolución R/01208/2014; DSK, 'Gesetzesentwurf zur Aufzeichnung von Fahrdaten ist völlig unzureichend!', 16 March 2017; ICO, 'Data Protection Technical Guidance Radio Frequency Identification' (2006); ICO, 'Wi-fi location analytics' (2016); Article 29-Data Protection Working Party (hereinafter ART29WP), 'Working document on data protection issues related to RFID technology', WP 105 (2005); ART29WP, 'Opinion 03/2017 on Processing personal data in the context of Cooperative Intelligent Transport Systems (C-ITS)', WP 252 (2017); ART29WP, 'Opinion 13/2011 on Geolocation services on smart mobile devices', WP 185 (2011).

[70] See GPDP, 9 July 2020, doc. web n. 9474649; GPDP, 4 December 2019, doc. web. n. 9215890; ICO (fn 68), Part. 3; CPVP, avis n. 10/2000, 3 April 2000; BFDI, 'Internet- und E-Mail Nutzung am Arbeitsplatz'; ART29WP, 'Working document on the surveillance of electronic communications in the workplace', WP 55 (2002).

[71] See GPDP, 4 December 2008, doc. web n. 1576125; GPDP, 24 February 2010, doc. web n. 1705070 (use of written coupons to authorise workers to leave their workstation to go to the toilet); AEPD, expediente n. E/01760/2017; AEPD, expediente n. E/01769/2017; CNIL, décision n. 2013-029, 12 July 2013; BFDI (fn 68); ICO, 'Installing CCTV? Things you need to do first'; ICO, 'In the picture: A data protection code of practice for surveillance cameras and personal information' (2017); ICO (fn 68), Part. 3; ICO, 'Wi-fi location analytics' (2016); ART29WP, 'Opinion 4/2004 on the Processing of Personal Data by means of Video Surveillance', WP 89 (2004). See also AEPD, procedimiento n. A/00109/2017, on the use of video surveillance systems by a hotel to monitor customers in relaxation areas.

[72] See GPDP, 21 July 2011, doc. web. n. 1825852; ICO (fn 68), Parts 1 and 4; Hamburg Commissioner for Data Protection and Freedom of Information, 1 October 2020 <https://datenschutz-hamburg.de/assets/pdf/2020-10-01-press-release-h+m-fine.pdf>, accessed 20 January 2021. See also GPDP, 11 January 2007, doc. web n. 1381620 (collection of sensitive information (e.g. sexual habits) by a real estate agency).

Finally, DPAs pointed out how human dignity can also be affected by public disclosure of personal information, such as evaluation judgements (e.g. publication of exam results by schools[75] or employee evaluation ratings;[76] use of services of the so-called reputation economy[77]) or personal debt situations,[78] which may cause distress and embarrassment to individuals.

*4.2.2 Freedom from discrimination*

According to DPAs, discriminatory practices[79] may occur in many contexts and in relation to different types of personal data processing. Negative consequences may result, for example, from automated decision-making and profiling activities, which may perpetuate existing stereotypes and social segregation.[80]

With regard to AI-related applications, DPAs have focused on the risks of perpetuating discriminatory practices through automated profiling.[81] Moreover, as the criteria and functioning of algorithms are often opaque, individuals might not know that they are being profiled or not understand the potential consequences. In this context, DPAs have considered, *inter alia*, the risk of bias that may arise from online behavioural advertising[82] or IoT-based profiling.[83] Likewise, DPAs have referred to the use of data-intensive systems in the context of police services and law enforcement, such as predictive policing.[84]

Adverse discriminatory impacts may also result from the use of sensitive data to prevent or limit access to certain services or benefits. This is the case when sensitive information is requested by

---

[73] See GPDP, 1 August 2013, n. 384, doc. web n. 2578547; AEPD, Gabinete Jurídico, informe 0392/2011; CNIL n. 2008-492, 11 December 2008; CPVP, avis n. 17/2008, 9 April 2008; ART29WP, 'Opinion 3/2012 on developments in biometric technologies', WP193 (2012).

[74] See ART29WP, 'Opinion 8/2014 on the on Recent Developments on the Internet of Things', WP 223 (2014); ART29WP, WP 185 (fn 69). With regard to the collection of sensitive data likely to cause embarrassment and discomfort to the data subject, see also ART29WP, 'Opinion 2/2010 on online behavioural advertising', WP 171 (2010).

[75] See ICO, 'Publication of exam results by schools' (2014). In this regard, particular concerns were also expressed by ART29WP, 'Opinion 2/2009 on the protection of children's personal data (General Guidelines and the special case of schools)', WP 160 (2009); GPDP, 'Scuola: Privacy, pubblicazione voti online è invasiva. Ammissione non sull'albo ma in piattaforme che evitino rischi', doc. web n. 9367295 (2020).

[76] See GPDP, 13 December 2018, n. 500, doc. web n. 9068983; BFDI, 'Notenspiegel im Intranet'.

[77] For instance, platforms which display and manage product and service reviews, as well as tax or criminal information. See GPDP, 24 November 2016, n. 488, doc. web n. 5796783.

[78] See GPDP, 28 May 2015, n. 319, doc. web n. 4131145; AEPD, procedimiento n. A/00104/2017.

[79] See, FRA, 'Handbook on European non-discrimination law' <https://fra.europa.eu/en/publication/2011/handbook-european-non-discrimination-law-2011-edition>, (2011), 22-29, accessed 11 January 2021; ECtHR, 13 November 2007, D.H. and Others v. the Czech Republic, App. No. 57325/00, para. 175. With reference to the difference between direct and indirect discrimination see also Claire Kilpatrick, 'Art. 21 – Non-discrimination', in Peers et al. (fn 65), 592.

[80] See ICO, 'Big data, artificial intelligence, machine learning and data protection' (2017); ICO, 'Guidance on AI and data protection'; CNIL, 'Comment permettre à l'homme de garder la main ? Les enjeux éthiques des algorithmes et de l'intelligence artificielle' (2017); ART29WP, 'Guidelines on Automated individual decision-making and Profiling for the purposes of Regulation 2016/679', WP 251 (2017).

[81] See ICO, 'Big data, artificial intelligence, machine learning and data protection' (2017).

[82] See ART29WP, WP 171 (fn 69); ICO (fn 76).

[83] See ART29WP, WP 185 (fn 69).

[84] See DSK, 'Big Data zur Gefahrenabwehr und Strafverfolgung: Risiken und Nebenwirkungen beachten', 18-19 March 2015. See also GPDP, 'Uomini e Macchine. Protezione dati per un'etica del digitale', doc. web n. 7598686 (2018); ICO, 'Guidance on AI and data protection'.

the employer (e.g. medical information), during the period of employment[85] or even at the time of recruitment,[86] or by real estate agencies in order to meet the discriminatory requirements of property owners.[87] Sensitive data for discriminatory purposes may also be used by insurance companies, which may collect genetic data to calculate insurance costs on the basis of foreseeable individual health conditions.[88]

Finally, in the case of discrimination too, monitoring and video-surveillance systems can have a negative impact on individuals and groups. This has led DPAs to highlight the unlawfulness of surveillance based exclusively on, *inter alia*, racial origin, religious or political opinions, membership in trade unions, or sexual orientation, without a justified reason.[89]

*4.2.3 Physical, psychological, and social identity*

Both the international and European legal frameworks consider personal identity in the broader context of individual privacy.[90] The personal aspects of an individual's identity traditionally cover different dimensions – physical, psychological, and social identity – and a range of data (e.g. name, image, reputation, family and ethnic heritage, gender identification, sexual, political and religious orientation).[91]

The notion of personal identity can thus encompass two different meanings: (i) the body of information that unequivocally identifies a person, distinguishing him or her from any other; (ii) information concerning the individual's projection in the social community.

The documents examined considered only the first meaning, with regard to data processing operations for personal identification.

One example is the use of biometric data to control access to certain areas (e.g. preventing outsiders from entering schools, tracking employees' whereabouts)[92] or the use of genetic data.[93] The need to protect personal identity also emerges from the considerations expressed by the authorities in relation to the identification information collected in the social media environment[94] and through RFID systems.[95]

---

[85] See ICO (fn 68), Part. 4; ICO, *The employment Practices Code. Supplementary Guidance*, Part. 4, 2005; GPDP, 5 June 2019, n. 146, doc. web n. 9124510.
[86] See ICO (fn 68), Part. 1; CNIL, 'Les opérations de recrutement' (2013); GPDP, 5 June 2019, n. 146, doc. web n. 9124510; ART29WP, WP 91, 'Working Document on Genetic Data' (2004).
[87] See GPDP, 11 January 2007, doc. web n. 1381620; CPVP, recommandation n. 01/2009, 18 March 2009.
[88] See ART29WP, WP 91 (fn 81). See also ART29WP, WP193 (fn 68) (on the potential discriminatory effects that may arise from the use of biometric data for targeting and profiling purposes).
[89] See ART29WP, WP 89 (fn 71).
[90] See, article 12 UDHR; article 7 EUCFR; article 8 ECHR.
[91] See Jens Vested-Hansen, 'Article 7 – Respect for Private and Family Life (Private Life, Home and Communications)', in Peers et al. (fn 65), 161; Ivana Roagna, 'Protecting the right to respect for private and family life under the European Convention on Human Rights' (Council of Europe, Strasbourg 2012), 12 <https://rm.coe.int/16806f1554> accessed 2 February 2021; Gert Brüggemeier, Aurelia Colombi Ciacchi and Patrick O'Callaghan, 'A common core of personality protection', in Gert Brüggemeier, Aurelia Colombi Ciacchi and Patrick O'Callaghan, Personality Rights in European Tort Law (Cambridge University Press 2010) 573-574. See also ECtHR, 20 July 2010 (Final 20 October 2010), Dadouch v. Malta, App. No. 38816/07, paras 47-48.
[92] See, among others, CNIL n. 2016-017, 28 January 2016; GPDP, 15 June 2006, doc. web n. 1306098; AEPD, Gabinet Juridico, informe 0392/2011; ICO, 'The use of biometrics in schools' <https://schools.essex.gov.uk/data/information-governance/Documents/biometrics.pdf> accessed 17 February 2021. See also CPVP, avis n. 17/2008, 9 April 2008; ART29WP, WP193 (fn 68); GPDP, 18 June 2015, n. 360, doc. web n. 4170232 (use of a facial recognition system by a company operating in the cruise travel sector).
[93] See ART29WP, WP 91 (fn 86).
[94] See, Art. 29WP, Opinion 5/2009 on online social networking, WP163 (2009).
[95] See GPDP, 9 March 2005, doc. web n. 1109493; ART29WP, WP 105 (fn 69).

In the context of personal identity, the potential unauthorised or unlawful use of identification information is of particular concern to the DPAs as this information cannot be modified (biometric and genetic data) or cannot be easily modified. Given the importance of this information, the DPAs have emphasised the need to limit its use and to ensure an adequate level of security to prevent identity theft and other offences.

*4.2.4 Physical, psychological and moral integrity and the intimate sphere*

Personal integrity is protected at European and international level as an aspect of an individual's private life,[96] and comprises the individual's physical, psychological and moral integrity.[97] In this sense, a natural person must be free from any interference, both in relation to the body and the mind. Respect for the intimate sphere of the data subject is also an important aspect of safeguarding individuals' integrity, referring to its moral dimension.[98]

The importance of the individual's physical integrity and intimate sphere is confirmed in DPA jurisprudence and opinions. Regarding the data subject's physical integrity, the DPAs mainly considered invasive data processing, such as that using implanted RFID devices (e.g. subcutaneous microchips) to collect and process personal information, including identification data, credit card number or health information.[99] The DPAs limit their use to situations where they are strictly necessary and there are no less intrusive alternatives, giving data subject the right to ask for their removal at any time.

With regard to the individual's intimate sphere, DPAs have focused on monitoring tools, including video-surveillance in environments where privacy expectations are high,[100] as well as on the collection of biometric data, given their invasive nature,[101] which could interfere with the data subject's intimate sphere in an excessive way.

The last group of cases concerns data processing operations carried out using IoT wearable devices or other devices used in close vicinity to the human body in daily life (e.g. smartphones and smart home devices).[102] The DPAs pointed out that the use of such devices is likely to cause significant interference to the individual's intimate sphere, by gathering information on health condition, behaviour, location, intimacy and many other aspects of the data subject.

---

[96] See article 12 UDHR; article 7 EUCFR ; article 8 ECHR. See also ECtHR, 24 July 2012 (Final 24 October 2012), Dordevic v. Croatia, App. No. 41526/10, para 97; ECtHR, 26 March 1985, X and Y v. Netherlands, App. No. 8978/80, para 22; ECtHR, 22 October 1996, Stubbings and Others v. the United Kingdom, App. No. 22083/93 and 22095/93, para 61. Serious matters relating to physical and mental integrity fall under Articles 3 EUCFR and 3 ECHR. See also Roagna (fn 86), 24; Sabine Michalowski, 'Article 3 - Right to the Integrity of the Person' in Peers et al. (fn 65), 42.

[97] See Lars Adam Rehof, 'Article 12', in Asbjørn et al. (fn 64), 187; Roagna (fn 86), 12; Vested-Hansen(fn 91), 156.

[98] See ECtHR, 28 May 2015, Y. v. Slovenia, App. No. 41107/10.

[99] See AEPD, Gabinete Jurídico, informe 0292/2010; GPDP, 9 March 2005, doc. web n. 1109493.

[100] Several cases concern restrooms and changing rooms and other places where privacy expectations are high: ICO, 'In the picture: A data protection code of practice for surveillance cameras and personal information' (2017); ICO (fn 68), Part. 3; ICO, 'Wi-fi location analytics' (2016); CNIL n. 2014-307, 17 July 2014; CNIL, décision n. 2013-029, 12 July 2013; GPDP, 10 July 2014, doc. web n. 3325380; AEPD, procedimiento n. A/00109/2017; ART29WP, WP 160 (fn 75); ART29WP, WP 89 (fn 71).

[101] See, among others, CNIL n. 2008-492, 11 December 2008; GPDP, 31 January 2013, doc. web n. 2304669; GPDP, 30 May 2013, doc. web n. 2502951; AEPD, Gabinete Jurídico, informe 0065/2015; ART29WP, WP193 (fn 68).

*4.2.5 Self-determination and personal autonomy*

Individual self-determination and personal autonomy are widely recognised at both international and European level.[103] Personal autonomy is protected as an aspect of individual private life[104] and safeguards individuals against a wide range of external interference.[105]

Individual self-determination and personal autonomy necessarily entail the ability to freely take decisions and have them respected by others[106]. According to international and European human rights jurisprudence, individual personal autonomy also covers a further range of human behaviours, among which are the right to develop one's own personality and the right to establish and develop relationships with other people[107], the right to pursue one's own aspirations and to control one's own information.[108]

Individual self-determination and personal autonomy also represent foundational principles in data protection, which is why it is not surprising that DPAs often refer to them in a broad sense.[109] These aspects emerge in both individual and relational contexts and involve freedom of choice, including freedom of movement and action, the free development of human personality and the right to informational self-determination.

Regarding freedom of choice, which encompasses freedom of movement and action, the DPAs have paid particular attention to the possible adverse effects of continuous and invasive monitoring. For instance, they considered cases of data processing carried out using video surveillance systems in workplaces[110] and schools[111] or in public spaces[112] (e.g. through the use of drones). The authorities also focus attention on the potentially negative outcomes arising from the use of devices such as wearable devices and smart meters.[113] Particular concerns were expressed in relation to monitoring activities through the use of mobile applications[114], as mobiles are strictly personal and are almost always on. Similarly, the authorities considered systems that collect mobility data, such as GPS, Wi-Fi tracking devices and RFID technologies.[115]

---

[102] See ART29WP, WP 223 (fn 74); ART29WP, WP 183, 'Opinion 12/2011 on smart metering' (2011); ICO, 'Privacy in mobile apps. Guidance for app developers' (2013).
[103] See, in particular: article 12 UDHR; article 7 EUCFR; article 8 ECHR.
[104] See, among others, Roagna (fn 86), 12; Vested-Hansen (fn 91), 156. See also ECtHR, 29 April 2002 (Final 29 July 2002), Pretty v. the United Kingdom, App. No. 2346/02, para 61.
[105] The individual personal autonomy is protected both in the private (e.g. home and workplace) and in the public context (e.g. against interferences from public authorities). See Rehof (fn 92), 187-201.
[106] See Commissioner for Human Rights, Human Rights and Disability, 'Equal rights for all', Strasbourg, 20 October 2008, para. 5.2. <https://rm.coe.int/16806dabe6> accessed 5 June 2018. Moreover, this interest is also protected in relation to communications; see Council of Europe, 'Guide on Article 8 of the European Convention on Human Rights. Right to respect for private and family life, home and correspondence', updated on 31 August 2019 <https://www.echr.coe.int/Documents/Guide_Art_8_ENG.pdf>, accessed 13 January 2021.
[107] See ECtHR, 18 May 1976, X v. Iceland, App. No. 6825/74.
[108] See Eva Fialová, 'Data Portability and Informational Self-Determination', in Masaryk University Journal of Law and Technology, 2014, (8), 45-55; Edward J. Eberle, 'The Right to Information Self-Determination' (2001) 4 Utah L. Rev. 965-1016.
[109] See also Tom L. Beauchamp, James F. Childress, *Principles of Biomedical Ethics* (Oxford University Press 2001), 63 ("to respect an autonomous agent is, at a minimum, to acknowledge that person's right to hold views, to make choices, and to take actions based on personal values and beliefs").
[110] See CNIL n. 2010-112, 22 April 2010; GPDP, 30 October 2013, n. 484, doc. web n. 2908871; ICO (fn 68), Part. 3; BFDI (fn 68); ART29WP, 'Opinion 2/2017 on data processing at work', WP 249 (2017); ART29WP, WP 89 (fn 71).
[111] GPDP, 8 May 2013, n. 230, doc. web n. 2433401; CPVP (fn 63), avis, n. 8/2006 del 12 April 2006; ART29WP, WP 160 (fn 75).
[112] CNIL, deliberation n. 94-056, 21 June 1994; ICO, 'CCTV code of practice. Draft for consultation' (2014); DSK, 'Einsatz von Videokameras zur biometrischen Gesichtserkennung birgt erhebliche Risiken', 30 March 2017; DSK (fn 63); ART29WP, WP 231, 'Opinion 01/2015 on Privacy and Data Protection Issues relating to the Utilisation of Drones' (2015); ART29WP, WP 89, (fn 71). See also DSK (fn 64).

Individual freedom of choice could be also undermined by communications monitoring,[116] which might limit and influence individuals with respect to content and the decision to communicate it. The DPAs clearly recognise that all these activities can limit or influence individual self-determination. Constant monitoring can have adverse consequences in terms of kerbing the data subject's behaviour in such a way as to comply with the controller's wishes[117]. Invasive monitoring can be more serious depending on the context in which it is carried out. This is the case in the workplace where there is an imbalance of power between the employer and the employee.[118]

Self-determination and autonomy can also be affected by the aforementioned services of the so-called 'reputation economy'[119], which limit or influence the choices and behaviours of those who want to avoid negative opinions, and by profiling,[120] which can lock data subjects into a specific category and restrict them to their suggested preferences.[121]

From a different perspective, DPAs use the broad notion of individual self-determination and autonomy to safeguard the free and full development of individual personality and the right to establish and develop relationships with other people. This happens with regard to monitoring communications and online behaviour[122] or the use of video-surveillance systems[123], as they can affect the data subject's freedom to establish and develop relationships with other people. [124] Similarly, negative consequences may also derive from data processing operations involving special categories of individuals, such as minors, or sensitive data. Here, for example, the DPAs mention the publication of exam results by schools, which can cause embarrassment and inhibit free relationships among students, [125] and the publication by the media of information concerning children who are victims of violence and abuse. [126]

---

[113] See ART29WP, WP 223 (fn 74); ART29WP, WP 183 (fn 97).

[114] See ICO (fn 97).

[115] With reference to the monitoring outside the workplace, see CNIL n. 2014-294, 22 July 2014; CNIL n. 2005-278, 17 November 2005; GPDP, 7 November 2013, n. 499, doc. web n. 2911484; GPDP, 9 March 2005, doc. web n. 1109493; CPVP, avis n. 27/2009, 28 October 2009; CPVP, recommandation n. 01/2010, 17 March 2010; DSK, 'Keine PKW-Maut auf Kosten des Datenschutzes!', 14 November 2014; DSK, 'Datenschutz im Kraftfahrzeug – Automobilindustrie ist gefordert', 8-9 October 2014; ICO, 'Data Protection Technical Guidance Radio Frequency Identification' (2006); ICO, 'Wi-fi location analytics' (2016); ART29WP, 'Working Party 29 Opinion on the use of location data with a view to providing value-added services', WP 115 (2005); ART29WP, WP 185 (fn 69); ART29WP, WP 252 (fn 69); ART29WP, WP 105 (fn 69). With regard to the collection of mobility data within the work context see GPDP, 28 June 2018, n. 396, doc. web n. 9023246; GPDP, 18 May 2016, n. 226, doc. web n. 5217175; CPVP, avis n. 12/2005, 7 September 2005; CPVP, recommandation n. 03/2013, 24 April 2013; CNIL n. 2013-366, 23 November 2013; CNIL n. 2006-066, 16 March 2006; ICO (fn 68), Part. 3.

[116] See GPDP, 1 February 2018, doc. web n. 8159221; GPDP, 8 March 2018, n. 139, doc. web n. 8163433; AEPD, Gabinete Jurídico, informe 0464/2013; CPVP, avis n. 39/2001, 8 October 2001; ICO (fn 68), Part. 3; ICO, 'The employment Practices Code. Supplementary Guidance' (2005), Part. 3; BFDI (fn 68); ART29WP, WP 55 (fn 70); ART29WP, 'Opinion 04/2014 on surveillance of electronic communications for intelligence and national security purposes', WP 215 (2014).

[117] See, among others, Jonathan W. Penney, 'Chilling Effects: Online Surveillance and Wikipedia Use' (2016) 31(1) BTLJ 117-182; Roger Clarke, 'The regulation of civilian drones' impacts on behavioural privacy' in this Review, 2014, 30(3), 263-285; Daniel J. Solove, 'A Taxonomy of Privacy' (2006) 154(3) U. Pa. L. Rev. 477-560.

[118] DPAs have identified a set of limits to employers' monitoring, see, among others, GPDP, 24 February 2010, doc. web n. 1705070; CNIL n. 2009-201, 16 April 2009; ICO (fn 68), Part. 3; CPVP, avis, n. 08/2006, 12 April 2006.

[119] See GPDP, 24 November 2016, n. 488, doc. web n. 5796783.

[120] See ART29WP, WP 251 (fn 80).

[121] See also Eli Pariser, *The filter bubble. What the Internet is Hiding from You* (Penguin books 2011).

Finally, DPAs consider the role of individual self-determination and autonomy to safeguard the data subject's right to informational self-determination and in order to protect use of their own personal data. This is the case, for example, of mandatory consent to access services or to access them on more advantageous terms (e.g. access to social networks, [127] access to certain services and features of IoT devices[128], transport services[129] or energy services[130]).[131]

*4.2.6 Freedom of expression and freedom of thought, conscience and religion*

As in European and international legal systems[132], the DPAs take freedom of expression to include various elements, such as the freedom to hold an opinion[133], or to impart and receive information and ideas.[134]

---

[122] See GPDP, 4 December 2019, doc. web. n. 9215890; GPDP, 4 June 2015, n. 345, doc. web n. 4211000; GPDP, 2 February 2006, doc. web n. 1229854; ICO (fn 68), Part. 3.
[123] See ART29WP, WP 160 (fn 75).
[124] With reference to the protection of this right in the work context: see ECtHR, 16 December 1992, Niemietz v. Germany, App. No. 13710/88, para 29; ECtHR, 5 September 2017, Bărbulescu v. Romania, App. No. 61496/08, paras 70-73.
[125] See ICO (fn 70); See ART29WP, WP 160 (fn 75). On the same grounds the DPAs considered the disclosure of employees' level of performance and evaluation marks, see BFDI (fn 76).
[126] See GPDP, 10 July 2008, doc. web n. 1536583. See also GPDP, 15 November 2001, doc. web n. 39596.
[127] See ART29WP, 'Opinion 15/2011 on the definition of consent', WP 187 (2011).
[128] ART29WP, WP 223 (fn 74).
[129] CNIL n. 2009-002, 20 January 2009.
[130] See GPDP, 27 October 2016, n. 439, doc. web n. 5687770. With reference to compulsory consent in the context of other services see, for example, DSK, 'Novellierung des Personalausweisgesetzes - Änderungen müssen bürger- und datenschutzfreundlich realisiert werden!', 24 January 2017; DSK, 'Wearables und Gesundheits-Apps – Sensible Gesundheitsdaten effektiv schützen!', 6-7 April 2016; GPDP, 20 July 2017, doc. web n. 6955363; GPDP, 13 May 2015 n. 291, doc. web n. 4337465; ICO, 'Direct marketing' (2018).
[131] Negative consequences may also occur when consent is provided in a situation of power imbalance, such as in the workplace. See CPVP, avis, n. 08/2006, 12 April 2006. See also DSK, 'Wearables und Gesundheits-Apps – Sensible Gesundheitsdaten effektiv schützen!' (fn 130); ICO (fn 68), Part. 3; ART29WP, WP 187 (fn 127); ART29WP, 'Opinion 8/2001 on the processing of personal data in the employment context', WP 48 (2001).
[132] See, among others, Dominika Bychawska-Siniarska, 'Protecting the right to freedom of expression under the European Convention on Human Rights' (Council of Europe, Strasbourg 2017); Lauri Hannikainen and Kristian Myntti, 'Article 19', in Asbjørn et al. (fn 64), 275; Lorna Woods, 'Article 11 – Freedom of Expression and Information' in Peers et al. (fn 65), 311.
[133] The freedom to hold an opinion concerns the *forum internum* of a person and protects their thoughts. Individuals must not be indoctrinated by states or other actors. Promoting one-sided information can also be an unacceptable obstacle to the freedom to hold opinions.
[134] Individuals have the freedom to express information and ideas as well as the right to disseminate them. They also have the right to receive any information, opinion, report, or news made public. According to the courts, individual freedom to receive information and ideas also includes the right to be adequately informed, in particular on matters of public interest. Access to the Internet is seen as included in the freedom of expression by the courts, as a key vehicle for the transmission and reception of information and ideas, see ECtHR, 1 December 2015 (Final 1 March 2016), Cengiz and Others v. Turkey, App. No. 48226/10 and 14027/11.

According to the DPAs, data subjects' freedom of expression may be constrained, for example, by use of targeted AI systems in political campaigns,[135] to influence voters and manipulate outcomes. Similarly, the DPAs pay attention to the interference with freedom that may occur when social networks automatically block access to a political group's page on the grounds of unverified complaints[136]. Moreover, the DPAs stress the need to safeguard the individual's freedom of expression in relation to fake news and online disinformation, underlining how misleading or false information may influence the public's political opinions.[137]

Negative consequences for freedom of expression and, in particular, for the freedom to receive information, may also arise from the publication of incorrect, obsolete information,[138] or unreal news by media.[139] In the same way, the DPAs consider individual freedom to receive information in assessing the legitimacy of data subjects' requests to remove or conceal information relating to them because they were unlawfully acquired.[140]

Finally, the DPAs take into account the potential prejudice to the data subject's freedom of thought, conscience and religion[141] that may derive from certain data processing operations. This is the case, for example, of the use of video surveillance systems in places of worship without security purposes or alternative measures,[142] as this may condition and limit individuals' activity.

*4.2.7 Freedom of assembly and association*

Though only in a limited number of decisions, DPAs also consider the need to safeguard the data subject's freedom of assembly and association.[143] They recognise that negative outcomes for the data subject's freedom of assembly may result for example from the gathering of identification data of participants in a trade union rally,[144] as this may discourage some from taking part in it.[145] Similarly, negative effects on individuals' freedom of assembly may arise from the use of drones by the police and other law enforcement authorities to monitor public demonstrations or similar gatherings.[146]

---

[135] See ICO, 'Democracy disrupted? Personal Information and political influence' (2018). See also GPDP, 18 April 2019, doc. web n. 9105201; CNIL n. 2012-021, 21 January 2012.

[136] See GPDP, 26 November 2019, doc. web n. 9195349. See also GPDP, 16 September 2019, doc. web n. 9138934.

[137] See also EDPS, 'Opinion 3/2018 EDPS opinion on online manipulation and personal data' (2018) <https://edps.europa.eu/sites/edp/files/publication/18-03-19_online_manipulation_en.pdf> accessed 18 January 2021; Konferenz der Informationsfreiheitsbeauftragten, 'Mit Transparenz gegen „Fake-News"', 13 June 2017.

[138] See also ECtHR, 16 July 2013 (Final 16 October 2013), Węgrzynowski and Smolczewski v. Poland, App. No. 33846/07 (Internet archives were considered to be covered by freedom of expression, as a fundamental source for education and research).

[139] See GPDP 15 September 2016, doc. web n. 5515910; GPDP 24 January 2013, doc. web n. 2286820; Autorité de protection des données (Belgian DPA, former Commission de la protection de la vie privée, CPVP), 14 July 2020 (n. DOS-2019-03780); ART29WP, 'Guidelines on the implementation of the Court of Justice of the European Union Judgment on "Google Spain and INC V. Agencia Española de Protección de datos (AEPD) and Mario Costeja Gonzàlez" C-131/12' (2014).

[140] See, among others, GPDP, 26 November 2020, doc. web n. 9509558; GPDP, 10 July 2014, doc. web n. 3352396.

[141] See article 9 ECHR; article 10 EUCFR; article 18 UDHR. The right to freedom of thought, conscience and religion protects an individual's fundamental beliefs and the right to manifest those beliefs both individually and with others, in both the private and public sphere. Non-religious viewpoints are also taken into consideration. Thus, the imposition upon individual actions or practices contrary to personal beliefs, as well as restrictions on individual actions or behaviours imposed by belief, will fall within the scope of the guarantee. See Jim Murdoch, *Protecting the right to freedom of thought, conscience and religion under the European Convention on Human Rights* (Council of Europe, Strasbourg, 2012) 7. See also Ronan McCrea, 'Art. 10 – Right to Freedom of Thought, Conscience and Religion', in Peers et al. (fn 65), 300-302; Martin Scheinin, 'Article 18' in Asbjørn et al. (fn 64), 263-274.

[142] AEPD, expediente n. E/03614/2017; GPDP, 23 February 2017, doc. web n. 6040861. See also ICO, 'Wi-fi location analytics' (2016).

[143] See Council of Europe, 'Guide on Article 11 of the European Convention on Human Rights' (2020), <https://www.echr.coe.int/Documents/Guide_Art_11_ENG.pdf>, accessed 10 February 2021.

*4.2.8 The right to the confidentiality of communications*

According to DPA jurisprudence, this right[147] is relevant, for example, in cases of firms monitoring their employees' electronic communications (telephone conversations, e-mails, and social media[148]), where the consequences of any breach of confidentiality might affect not only the workers, but also others, such as the worker's family members and the company's customers. Here the authorities stress the need to balance the workers' right to secrecy of correspondence (and that of the other individual involved) with the legitimate rights and interests of the employer.

Furthermore, the DPAs consider the right to confidential communications in relation to monitoring electronic communications and traffic data retention for national security purposes, where any interference with this fundamental right is allowed only if it is strictly necessary in the interests of national security.[149]

**5. A proposal for an HRIA model**

The analysis described in Section 4 and the evidence provided by DPAs' decisions and practice show that the issues concerning the use of data-intensive applications are not circumscribed to the debated topic of bias and discrimination but have a broader impact on several human rights and freedoms. For this reason, a comprehensive HRIA model is needed.

It is worth noting that traditional HRIAs are often territory-based considering the impact of business activities in a given local area and community, whereas in the case of AI applications this link with a territorial context may be less significant.

There are two different scenarios: cases characterised by use of AI in territorial contexts with a high-impact on social dynamics (e.g. smart cities plans, regional smart mobility plans, predictive crime programmes) and those where AI solutions have a more limited impact as they are embedded in globally distributed products/services (e.g. AI virtual assistants, autonomous cars, recruiting AI-based software, etc.) and do not focus on a given socio-territorial community. While in the first case the context is very close to the traditional HRIA cases, where large-scale projects affect whole communities and the potential impacts cover a wide range of human rights, the second case is characterised by a more limited social impact, centred more on individuals rather than on society at large.[150] This difference has a direct effect on the structure and complexity of the model, as well as the tool employed.

---

[144] GPDP, 29 November 2012, doc. web n. 2192643.
[145] See also ECtHR, 14 February 2006 (Final 14 May 2006), Christian Democratic People's Party v. Moldova, App. No. 28793/02, para 77.
[146] ART29WP, WP 231 (fn 112), 11.
[147] The right to respect for communications is protected in the international and European regulatory framework as an important aspect of private life, see Article 12 UDHR, Article 7 EUCFR, and Article 8 ECHR. In this context, the protection of communications includes not only correspondence of a personal nature but also that with professional and commercial content. See e.g ECtHR, 5 September 2017, Bărbulescu v. Romania (fn 124), paras 70-73; ECtHR, 22 May 2008 (Final 22 August 2008), Ilya Stefanov v. Bulgaria, App. No. 65755/01.
[148] GPDP, 1 February 2018, doc. web n. 8159221; GPDP, 4 June 2015, n. 345, doc. web n. 4211000; BFDI (fn 68); AEPD, Gabinete Jurídico, informe 0464/2013; CPVP, Recommandation n. 08/2012, 2 May 2012; CNIL, 'Le contrôle de l'utilisation d'internet et de la messagerie électronique' (2015); ICO (fn 68), Part. 3; ART29WP, WP 55 (fn 70).
[149] ART29WP, WP 215 (fn 116); ART29WP, WP 113, *Opinion 4/2005 on the Proposal for a Directive of the European Parliament and of the Council on the Retention of Data Processed in Connection with the Provision of Public Electronic Communication Services and Amending Directive 2002/58/EC (COM(2005)438 final of 21.09.2005)*, adopted on 21 October 2005.

Criteria such as the AAAQ framework[151], for example, or issues concerning property and lands, can be used in assessing a smart city plan, but are unnecessary or disproportionate in the case of an AI-based recruitment software. Similarly, a large-scale mobility plan may require a significant monitoring of needs through interviews of right holders and stakeholders, while in the case of an AI-based personal IoT device this phase can be much reduced.

Regarding the first and more complex scenarios, we provide only a limited contribution, as the existing HRIA models can be used in those cases. Here, a greater focus on data-intensive systems leads to a reflection on the challenges that large-scale poses with regard to multi-factor scenarios (Section 6.2).

We expect our model to make more significant contribution in the second scenario, where the traditional eighth/twelve-month HRIA should be scaled down to a more manageable size for small-scale projects, and focused on quantifiable criteria to be applied in AI product and service development.

The model described here consists of two main building blocks, which are examined in detail in the following sub-sections: planning and scoping, and data collection and analysis.

*5.1 Planning and scoping*

The first stage deals with definition of the HRIA target, identifying the main features of the product/service and the context in which it will be placed, in line with the context-dependent nature of the HRIA. Three are the main areas to consider at this stage:

- description and analysis of the type of product/service, including data flows and data processing purposes
- the human rights context (contextualisation on the basis of local jurisprudence and laws)
- identification of relevant stakeholders.

The table below provides a non-exhaustive list of potential questions for HRIA planning and scoping. The extent and content of these questions will depend on the specific nature of the product/service and the scale and complexity of its development and deployment.[152] This list is therefore likely to be further supplemented with project-specific questions.[153]

---

[150] This does not mean that the collective dimension does not plays an important role and should be adequately considered in the assessment process, see Alessandro Mantelero, 'Personal data for decisional purposes in the age of analytics: From an individual to a collective dimension of data protection' in this Review (2016), 32 (2), 238-255.
[151] See Danish Institute for Human Rights, 'The AAAQ Framework and the Right to Water: International indicators for availability, accessibility, acceptability and quality' (2014).
[152] See e.g. The Danish Institute for Human Rights, 'Scoping practitioner supplement. Human rights impact assessment guidance and toolbox' (Copenhagen 2020); The Danish Institute for Human Rights, 'Guidance on HRIA of Digital Activities. Phase 1: Planning and scoping' (Copenhagen 2020) <https://www.humanrights.dk/sites/humanrights.dk/files/media/document/Phase%201_%20Planning%20and%20Scoping_n.pdf> accessed 20 February 2021.
[153] For similar questionnaires, see e.g. p. 23-30 The Danish Institute for Human Rights, 'Human Rights Impact Assessment Guidance and Toolbox' (Copenhagen 2016)

Tab. 1. Planning & scoping

| Description and analysis of the type of product/service, including related data flows and data processing purposes | <ul><li>What are the main features of the product/service?</li><li>In which countries will the product/service be offered?</li><li>Identification of rights-holders: who are the target-users of the product/service?</li><li>What types of data are collected (personal, non-personal, special categories)?</li><li>What are the main purposes of data processing?</li><li>Identification of the duty-bearers: which subjects are involved in data management and what is their role in data processing?</li></ul> |
|---|---|
| Human rights context (contextualisation based on local jurisprudence and laws) | <ul><li>Which human rights are potentially affected by the product/service?</li><li>Which international/regional legal instruments have been implemented at an operational level?</li><li>Which are the most relevant courts or authoritative bodies in the field of human rights in the context?</li><li>What are the relevant decisions and provisions in the field of human rights?</li></ul> |
| Controls in place | <ul><li>What policies and procedures are in place to assess the potential impact on human rights, including stakeholder engagement?</li><li>Has an impact assessment been carried out, developed and implemented in relation to specific issues or some features of the product/service (e.g. use of biometrics)?</li></ul> |
| Stakeholder engagement | <ul><li>Which are the main groups or communities potentially affected by the service/product, including its development?</li><li>What other stakeholders should be involved, in addition to affected community and groups, (e.g. civil society and international originations, experts, industry associations, journalists)?</li><li>Are there any other duty-bearers to be involved, apart from the product/service developer (e.g. national authorities, governmental agencies)?</li><li>Were business partners, including suppliers (e.g. subcontractors in AI systems and datasets) involved in the assessment process?</li><li>Has the developer conducted an assessment of its supply chain to identify whether the activities of suppliers/contractors involved in product/service development might contribute to adverse human rights impacts? Has the developer promoted human rights standards or audits to ensure respect for human rights among suppliers?</li><li>Do the product/service developers publicly communicate the potential impacts on human rights of the service/product?</li></ul> |

---

<https://www.humanrights.dk/business/tools/human-rights-impact-assessment-guidance-and-toolbox> accessed 18 June 2020.

|  | • Does the developer provide training on human rights standards for relevant management and procurement staff? |
|---|---|

*5.2 Data collection and analysis*

While the first stage is mainly desk research, the second focuses on gathering relevant empirical evidence to assess the product/service's impact on human rights and freedoms. In traditional HRIA this usually involves extensive fieldwork. But in the case of AI applications, data collection and analysis is restricted to large-scale projects such as those developed in the context of smart cities, where different services are developed and integrated. For the remaining cases, given the limited and targeted nature of each application, data collection is largely related to the product/service's features and feedback from stakeholders.

Based on the information gathered in the previous stage (description and analysis of the type of product/service, human rights context, controls in place, and stakeholder engagement), we can proceed to a contextual assessment of the impact of data use on human rights, to understand which rights and freedoms may be affected, how this may occur, and which potential mitigation measures may be taken.

Since in most cases the assessment is not based on measurable variables, the impact on rights and freedoms is necessarily the result of expert evaluation,[154] where expert opinion relies on knowledge of case law, the literature, and the legal framework. This means that it is not possible to provide precise measurement of the expected impacts but only an assessment in terms of range of risk (i.e. low, medium, high, or very high).

The benchmark for this assessment is the evidence-based analysis using the methodology described in Section 4 and the results. Thus, different rights and freedoms may be relevant depending on the specific nature of the given application.

Examination of any potentially adverse impact should begin with a general overview followed by a more granular analysis where the impact is envisaged.[155] In line with normal risk assessment

---

[154] See e.g. Martin Scheinin and Helga Molbæk-Steensig, 'Pandemics and human rights: three perspectives on human rights assessment of strategies against COVID-19' (2021) <https://cadmus.eui.eu//handle/1814/69576> accessed 25 February 2021.

[155] For an analytical description of the main components of impact analysis, based on the experience in the field of data protection, see Heleen L Janssen, 'An approach for a fundamental rights impact assessment to automated decision-making' (2020) 10(1) International Data Privacy Law 76–106, which uses four benchmarks covering the traditional areas of risk analysis in the law (impacted rights, risks at design stages and during operation, balancing risks and interests, control and agency over data processing). As for the risk assessment, the model proposed by the author does not provide a methodology to combine the different elements of impact assessment or to estimate the overall impact. Moreover, the model is used for an ex post comparative analysis, rather than for iterative design-based product/service development, as does the model we present here. In this sense, by providing two fictitious basic cases, Janssen tests her model though a comparative analysis (one case against the other) and without a clear analysis of the different risk components, in terms of individual impact and probability, with regard to each potentially affected right or freedom (e.g. "given that the monitor sensor captures every noise in its vicinity in situation (1), it probably has a high impact on a number of privacy rights, including that of intimacy of the home, communication privacy and chilling effects on the freedom of speech of (other) dwellers in the home"), and without a clear description of the assessment of their cumulative effect and overall impact. With a focus on the GDPR, see Margot E Kaminski and Gianclaudio Malgieri, 'Algorithmic Impact Assessments under the GDPR: Producing Multi-Layered Explanations' (2020) International Data Privacy Law DOI: 10.1093/idpl/ipaa020. See also Dillon

procedures, three key factors must be considered: risk identification, likelihood (L), and severity (S). As regards the first, the focus on human rights and freedoms already defines the potentially affected categories and the case specific analysis identifies those concretely affected, depending on the technologies used and their purposes. Since this is a rights-based model, risk concerns the prejudice to rights and freedoms, in terms of unlawful limitations and restrictions, regardless of material damage.

The expected impact of the identified risks is assessed by considering both the likelihood and the severity of the expected consequences, using a four-step scale (low, medium, high, very high) to avoid any risk of average positioning.

Likelihood is the combination of two elements: the probability of adverse consequences and the exposure. The former concerns the probability that adverse consequences of a certain risk might occur (Tab. 1) and the latter the potential number of people at risk (Tab. 2). In considering the potential impact on human rights, it is important not only to consider the probability of the impact, but also its extension in terms of potentially affected people.

Both these variables must be assessed on a contextual basis, considering the nature and features of the product and service, the application scenario, previous similar cases and applications, and any measures taken to prevent adverse consequences. Here, the engagement of relevant shareholders can help to better understand and contextualise these aspects, alongside the expertise of those carrying out the impact assessment.

Tab. 1 Probability

|  | Probability |  |
|---|---|---|
| Low | The risk of prejudice is improbable or highly improbable | 1 |
| Medium | The risk may occur | 2 |
| High | There is a high probability that the risk occurs | 3 |
| Very high | The risk is highly likely to occur | 4 |

Tab. 2 Exposure

|  | Exposure |  |
|---|---|---|
| Low | Few or very few of the identified population of rights-holders are potentially affected | 1 |
| Medium | Some of the identified population are potentially affected | 2 |
| High | The majority of the identified population is potentially affected | 3 |
| Very high | Almost the entire identified population is potentially affected | 4 |

These two variables are combined in the combinatorial table below using a cardinal scale to estimate the overall probability level (L). This table can be further modified on the basis of the context-specific nature of assessed AI systems and feedback received from experts and stakeholders.

Tab. 3 Likelihood table(L)

|  |  | Probability | | | |
|---|---|---|---|---|---|
|  |  | 1 | 2 | 3 | 4 |
| Expo | 1 | 1 | 2 | 3 | 4 |
|  | 2 | 2 | 3 | 5 | 9 |

---

Reisman and others, 'Algorithmic Impact Assessments: A Practical Framework for Public Agency Accountability' (2018) <https://ainowinstitute.org/aiareport2018.pdf> accessed 29 June 2018.

| | | | | | | | Likelihood | |
|---|---|---|---|---|---|---|---|---|
| | | | | | | | Low | 1 |
| | | | | | | | Medium | 2 |
| | 3 | 3 | 5 | 9 | 12 | | High | 3 |
| | 4 | 4 | 7 | 12 | 15 | | Very high | 4 |

The severity of the expected consequences (S) is estimated by considering the nature of potential prejudice in the exercise of rights and freedoms and their consequences. This is done by considering the gravity of the prejudice (gravity), and the effort to overcome it and to reverse adverse effects (effort).

Tab. 4 Gravity of the prejudice

| | Gravity of the prejudice | |
|---|---|---|
| Low | Affected individuals and groups may encounter only minor prejudices in the exercise of their rights and freedoms. | 1 |
| Medium | Affected individuals and groups may encounter significant prejudices. | 2 |
| High | Affected individuals and groups may encounter serious prejudices. | 3 |
| Very high | Affected individuals and groups may encounter serious or even irreversible prejudices. | 4 |

Tab. 5 Effort to overcome the prejudice and to reverse adverse effects.

| | Effort | |
|---|---|---|
| Low | Suffered prejudice can be overcome without any problem (e.g. time spent amending information, annoyances, irritations, etc.) | 1 |
| Medium | Suffered prejudice can be overcome despite a few difficulties (e.g. extra costs, fear, lack of understanding, stress, minor physical ailments, etc.). | 2 |
| High | Suffered prejudice can be overcome albeit with serious difficulties (e.g. economic loss, property damage, worsening of health, etc.). | 3 |
| Very high | Suffered prejudice may not be overcome (e.g. long-term psychological or physical ailments, death, etc.). | 4 |

As in the case of likelihood, these two variables are combined in the table below using a cardinal scale to estimate the severity level (S).

Tab. 6 Severity table (S)

| | | Gravity | | | |
|---|---|---|---|---|---|
| | | 1 | 2 | 3 | 4 |
| Effort | 1 | 1 | 2 | 4 | 6 |
| | 2 | 2 | 3 | 5 | 8 |
| | 3 | 3 | 5 | 8 | 10 |
| | 4 | 5 | 8 | 10 | 12 |

| Severity | |
|---|---|
| Low | 1 |
| Medium | 2 |
| High | 3 |
| Very high | 4 |

A table (Tab. 7) for the overall assessment charts both variables – likelihood (L) and severity (S) of the expected consequences – against each envisaged risk to rights and freedoms (R1, R2,… Rn).

Tab. 7. Table of envisaged risks

|    | L | S | overall impact |
|----|---|---|----------------|
| R1 |   |   |                |
| R2 |   |   |                |
| …. |   |   |                |
| Rn |   |   |                |

The overall impact for each examined risk, taking into consideration the L and S values, is determined using a further table (Tab. 8). The colours represent the overall impact, which is very high in the dark red sector, high in the red sector, medium in the yellow sector and is low in the green sector.

Tab. 8. Overall risk impact table

|            |           | Severity [impacted right/freedom] | | | |
|------------|-----------|------|--------|------|-----------|
|            |           | Low  | Medium | High | Very high |
| **Likelihood** | Low       | green | yellow | red | red |
|            | Medium    | yellow | yellow | red | red |
|            | High      | yellow | red | red | dark red |
|            | Very high | yellow | red | dark red | dark red |

Once the potentially adverse impact has been assessed for each of the rights and freedoms considered, a radial graph is charted to represent the overall impact on them. This graph is then used to decide the priority of intervention in altering the characteristics of the product/service to reduce the expected adverse impacts.[156]

Graph 1. Radial graph (impact) example

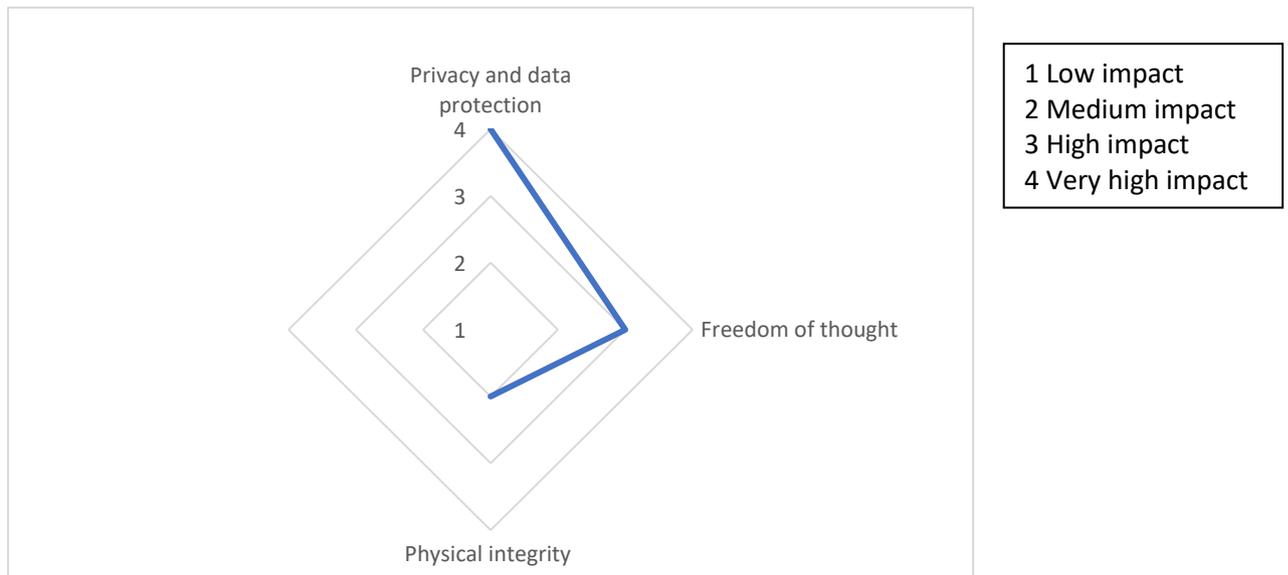

1 Low impact
2 Medium impact
3 High impact
4 Very high impact

To reduce the envisaged impacts, factors that can exclude the risk from a legal perspective (EFs) – such as the mandatory nature of certain impacting features or the prevalence of competing interests recognised by law – and those that can reduce the risk by means of appropriate mitigation measures (MMs) should be considered.

---
[156] This approach is also in line with the adoption of the Agile methodology in software development.

After the first adoption of the appropriate measures to mitigate the risk, further rounds of assessment can be conducted according to the level of residual risk and its acceptability, enriching the initial table with new columns (Tab. 9).

The first two new columns show any risk excluding factors (EFs) and mitigation measures (MMs), while the following two columns show the residual probability (rL) and severity (rS) of the expected consequences, after accounting for excluding and mitigation factors. The last column gives the final overall impact, using rL and rS values and the overall impact table (Tab 8); this result can also be represented in a new radial graph.

Tab. 9. Comparative risk impact analysis table (before/after mitigation measures and excluding factors)

|    | L | S | Overall impact | EFs | MMs | rL | rS | Final Impact |
|----|---|---|----------------|-----|-----|----|----|--------------|
| R1 |   |   |                |     |     |    |    |              |
| R2 |   |   |                |     |     |    |    |              |
| …. |   |   |                |     |     |    |    |              |
| Rn |   |   |                |     |     |    |    |              |

Note that it is possible to estimate overall impact, as an average of the impacts on all the areas analysed. But this necessarily treats all the different impacted areas (i.e. rights and freedoms) as having the same importance and is therefore a somewhat imprecise synthesis.

In terms of actual effects on operations, the radial graph is therefore the best tool to represent the outcome of the HRIA, showing graphically the changes after introducing mitigation measures. However, an estimation of overall impact could also be made in future since several legislative proposals on AI refer to an overall impact of each AI-based solution, using a single risk scale covering all potential consequences.

## 6. Testing the HRIA

The next two sub-sections examine two possible applications of the proposed model, with two different scales of data use. The first case, an Internet-connected doll equipped with AI, shows how the impact of AI is not limited to adverse effects on discrimination, but has a wider range of consequences (privacy and data protection, education, freedom of thought and diversity, etc.), given the innovative nature of the application and its interaction with humans.

This highlights the way in which AI does not merely concern data and data quality but more broadly the transformation of human-machine interaction by data-intensive systems. This is even more evident in the case of the smart cities, where the interaction is replicated on large scale affecting a whole variety of human behaviours by individuals, groups and communities.

The first case study (an AI-powered doll) shows in detail how the HRIA methodology can be applied in a real-life scenario. In the second case (a smart city project) we do not repeat the exercise for all the various data-intensive components, because a full HRIA would require extensive information collection, stakeholder engagement, and supply-chain analysis,[157] which go beyond the scope of this work.[158] But above all, the purpose of this second case study is different: to shed light on the dynamics of the HRIA in multi-factor scenarios where many different AI systems are combined.

Indeed, a smart city environment is not a single device, but encompasses a variety of technical solutions based on data and algorithms. The cumulative effect of integrating many layers results in a whole system that is greater and more complicated than the sum of its parts.

This explains why the assessment of potential risks to human rights and freedoms cannot be limited to a fragmented case-by-case analysis of each application. Rather, it requires an integrated approach that looks at the whole system and the interaction among its various components, which may have a wider impact than each component taken separately.

Scale and complexity, plus the dominant role of one or a few actors, can produce a cumulative effect which may entail multiple and increased impacts on rights and freedoms, requiring an additional integrated HRIA to give an overall assessment of the large-scale project and its impacts.

*6.1 Testing HRIA on a small scale: the Hello Barbie case*

Hello Barbie was an interactive doll produced by Mattel for the English-speaking market, equipped with speech recognition systems and AI-based learning features, operating as an IoT device. The doll was able to interact with users but did not interact with other IoT devices.[159]

The design goal was to provide a two-way conversation between the doll and the children playing with it, including capabilities that make the doll able to learn from this interaction, e.g. tailoring responses to the child's play history and remembering past conversations to suggest new games and topics.[160] The doll is no longer marketed by Mattel due to several concerns about system and device security.[161]

This section discusses the hypothetical case, imagining how the proposed assessment model[162] could have been used by manufactures and developers and the results that might have been achieved.

---

[157] See also Kate Crawford and Vladan Joler, 'Anatomy of an AI System: The Amazon Echo As An Anatomical Map of Human Labor, Data and Planetary Resources' (AI Now Institute and Share Lab 2018) <http://www.anatomyof.ai> accessed 27 December 2019.
[158] A proper HRIA would require a multidisciplinary team working locally for a significant period of time. For example, the human rights impact assessment of the Bisha Mine in Eritrea, which started in July 2013, issued its final HRIA report in February 2014, followed by an auditing procedure in 2015. See LKL International Consulting Inc., 'Human Rights Impact Assessment of the Bisha Mine in Eritrea' (2014) <https://media.business-humanrights.org/media/documents/files/documents/Nevsun_HRIA_Full_Report__April_2014_.pdf> accessed 26 October 2020; LKL International Consulting Inc., 'Human Rights Impact Assessment of the Bisha Mine in Eritrea 2015 Audit' (2015) <https://media.business-humanrights.org/media/documents/files/documents/Bisha-HRIA-Audit-2015.pdf> accessed 26 October 2020.
[159] See Mattel, 'Hello Barbie FAQ' Version 2 (2015) <http://hellobarbiefaq.mattel.com/faq/> accessed 12 November 2020.
[160] See 'Hello Barbie FAQ' (fn 159).
[161] See also Sharon Shasha et al., 'Playing With Danger: A Taxonomy and Evaluation of Threats to Smart Toys' (2019) 6(2) IEEE Internet of Things Journal 2986-3002 (with regard to Hello Barbie see Appendix A, para A.3).
[162] On the safeguard of human rights and the use of HRIA in the business context see United Nations, 'Guiding Principles on Business and Human Rights' (2011)

*6.1.1 Planning and scoping*

Starting with the questions listed in Tab. 1 above and information on the case examined, the planning and scoping phase would summarise the key product characteristics as follows:

a) A connected toy with four main features: (i) programmed with more than 8,000 lines of dialogue[163] hosted in the cloud, enabling the doll to talk with the user about "friends, school, dreams and fashion";[164] (ii) speech recognition technology[165] activated by a push-and-hold button on the doll's belt buckle; (iii) equipped with a microphone, speaker and two tri-colour LEOs embedded in the doll's necklace, which light up when the device is active; (iv) a Wi-Fi connection to provide for two-way conversation.[166]
b) The target-user is an English-speaking child (minor). Theoretically the product could be marketed worldwide in many countries, but the language barrier represents a limitation.
c) The right-holders can be divided into three categories: direct users (minors), supervisory users (parents, who have partial remote control over the doll and the doll/user interaction) and third parties (e.g. friends of the user or re-users of the doll).
d) Regarding data processing, the doll collects and stores voice-recording tracks based on dialogues between the doll and the user; this information may include personal data[167] and sensitive information.[168]
e) The main purpose of the data processing and AI is to create human-robot interaction (HRI) by using machine learning (ML) to build on the dialogue between the doll and its young users. There are also additional purposes: (i) educational; (ii) parental control and

---

<https://www.ohchr.org/Documents/Publications/GuidingPrinciplesBusinessHR_EN.pdf> accessed 8 December 2020 ("The State duty to protect is a standard of conduct. Therefore, States are not per se responsible for human rights abuse by private actors. However, States may breach their international human rights law obligations where such abuse can be attributed to them, or where they fail to take appropriate steps to prevent, investigate, punish and redress private actors' abuse") and more specifically Principles 13, 18 and 19. See also The Danish Institute for Human Rights (fn 153), 6-7.

[163] The comprehensive list of all the lines Hello Barbie says as of November 17, 2015 is available here: <http://hellobarbiefaq.mattel.com/wp-content/uploads/2015/11/hellobarbie-lines-v2.pdf> accessed 28 November 2020.

[164] 'Hello Barbie FAQ' (fn 159). Cloud service was provided by ToyTalk, see the following footnote.

[165] This technology and services were provided by ToyTalk, a Mattel's partner.

[166] See 'Hello Barbie FAQ' (fn 159).

[167] See 'Hello Barbie FAQ' (fn 159) ("Q: Can Hello Barbie say a child's name? No. Hello Barbie does not ask for a child's name and is not scripted to respond with a child's name, so she will not be able to recite a child's name back to them"). But see Meg Leta Jones, 'Your New Best Frenemy: Hello Barbie and Privacy Without Screens' (2016) 2 Engaging Science, Technology, and Society 242, 245 who reports this reply in the dialogue with the doll: "Barbie: Sometimes I get a little nervous when I tell people my middle name. But I'm really glad I told you! What's your middle name?".

[168] See 'Hello Barbie FAQ' (fn 159) ("Although Hello Barbie was designed not to ask questions which are intended to elicit answers that might contain personal information, we cannot control whether a child volunteers such information without prompting. Parents who are concerned about this can monitor their child's use of Hello Barbie, and parents have the power to review and delete any conversation their child has with Hello Barbie, whether the conversations contain personal information or not. If we become aware of any such personal information captured in recordings, it is our policy to delete such information, and we contractually require our Service Providers to do the same. This personal information is not used for any purpose").

surveillance[169] (parents can listen, store and re-use recorded conversations);[170] (iii) direct advertising to parents;[171] (iv) testing and service improvement.[172]

f) The chief duty-bearer is the producer, but in connected toys other partners – such as ToyTalk in this case – may be involved in the provision of ML, cloud and marketing services.

Another important set of data to be collected at this stage concerns the potential interplay with human rights and the reference framework, including main international/regional legal instruments, relevant courts or other authoritative bodies, and relevant decisions and provisions (see Tab. 1, the human rights context).

As regards the rights potentially affected, depending on the product's features and purposes, data protection and the right to privacy are the most relevant due to the possible content of the dialogue between the doll and the user, and the parental monitoring. Here the legal framework is represented by a variety of regulations at different levels. Compliance with the US COPPA[173] and the EU GDPR[174] can cover large parts of the potential market of this product and international guiding principles[175] can facilitate the adoption of global policies and solutions.

---

[169] See 'Hello Barbie FAQ' (fn 159) ("Hello Barbie only requires a parent's email address to set up an account. This is necessary so that parents can give permission to activate the speech recognition technology in the doll. Other information, such as a daughter's birthday, can be provided to help personalize the experience but are not required"). See also fn 165.

[170] See 'Hello Barbie FAQ' (fn 159) ("Hello Barbie recording and storing conversations girls have with the doll? Yes. Hello Barbie has conversations with girls, and these conversations are recorded. These audio recordings are used to understand what is being said to Hello Barbie so she can respond appropriately and also to improve speech recognition for children and to make the service better. These conversations are stored securely on ToyTalk's server infrastructure and parents have the power to listen to, share, and/or delete stored recordings any time").

[171] See 'Hello Barbie FAQ' (fn 159) ("Q. Are conversations used to market to children? No. The conversations captured by Hello Barbie will not be used to contact children or advertise to them." This was confirmed by the analysis carried out by Shasha et al. (fn 161). Regarding the advertising directs to parents, this is the answer provided in the FAQ: "Q: Your Privacy Policy says that you will use personal information to provide consumers with news and information about events, activities, promotions, special offers, etc. That sounds like consumers could be bombarded with marketing messages. Can parents elect not to receive those communications? Yes. Opting out of receiving promotional emails will be an option during the set up process and you can opt out at any time by following the instruction in those emails. Note that marketing messages will not be conveyed via the doll itself").

[172] See 'Hello Barbie FAQ' (fn 159) ("Conversations between Hello Barbie and consumers are not monitored in real time, and no person routinely reviews those conversations. Upon occasion a human may review certain conversations, such as in order to test, improve, or change the technology used in Hello Barbie, or due to support requests from parents. If in connection with such a review we come across a conversation that raises concern about the safety of a child or others, we will cooperate with law enforcement agencies and legal processes as required to do so or as we deem appropriate on a case-by-case basis").

[173] See Federal Trade Commission, 'Enforcement Policy Statement Regarding the Applicability of the COPPA Rule to the Collection and Use of Voice Recordings', 23 October 2017 <https://www.ftc.gov/public-statements/2017/10/federal-trade-commission-enforcement-policy-statement-regarding> accessed 28 November 2020. See also Eldar Haber, 'Toying with Privacy: Regulating the Internet of Toys' (2019) 80 Ohio State Law Journal 399.

[174] See also ICO, 'Age appropriate design: a code of practice for online services', section 14. Connected toys and devices (2020) < https://ico.org.uk/for-organisations/guide-to-data-protection/key-data-protection-themes/age-appropriate-design-a-code-of-practice-for-online-services/> accessed 20 February 2021.

[175] See e.g. Council of Europe, Convention 108+. See also Council of Europe, 'Recommendation CM/Rec(2018)7 of the Committee of Ministers. Guidelines to Respect, Protect and Fulfil the Rights of the Child in the Digital Environment' <https://rm.coe.int/guidelines-to-respect-protect-and-fulfil-the-rights-of-the-child-in-th/16808d881a> accessed 28 November 2020, para 36 ("With respect to connected

Moreover, in relation to data processing and individual freedom of choice, the potential effects of marketing strategies can also be considered as forms of freedom of expression[176] and freedom to conduct a business.

Given the broad interaction between the doll and the user and the behavioural, cultural and educational influence that the doll may have on young users,[177] further concerns relate to freedom of thought and diversity.[178]

In the event of cyberattack and data theft or transmission of inappropriate content to the user through the doll, safety issues also arise and may impact on the right to psychological and physical safety and health.

With the potentially global distribution of the toy, the possible impacts need to be further contextualised within each legal framework, taking into consideration local case law and that of regional supranational bodies like the European Court of Human rights. In this regard, it is necessary during the scoping phase to identify the significant provisions and decisions in the countries/regions where the product is distributed.

The last aspect to be considered in planning and scoping HRIA concerns the identification and engagement of potential stakeholders. In the case of connected toys, the most important stakeholders are likely to be parents' associations, educational bodies, professional associations (e.g. psychologists and educators), child, consumer and data protection supervisory bodies, as well as trade associations. Stakeholders may also include the suppliers involved in product/service development. In the latter case, the HRIA must also assess the activities by these suppliers and may benefit from an auditing procedure or the adoption of standards.

The following sections describe an iterative assessment process, starting from the basic idea of the connected AI-equipped toy with its pre-set functionality and moving on to a further

---

or smart devices, including those incorporated in toys and clothes, States should take particular care to ensure that data-protection principles, rules and rights are also respected when such products are directed principally at children or are likely to be regularly used by or in physical proximity to children"). See also Alessandro Mantelero, 'The future of data protection: Gold standard vs. global standard', in this Review, 2021, DOI: 10.1016/j.clsr.2020.105500.

[176] See Universal Declaration of Human Rights, article 19, and International Covenant on Civil and Political Rights, article 19(2). See also Human Rights Committee, General Comment no. 34 (CCPR/C/GC/34), para. 11; UNICEF and others, 'Children's Rights and Business Principles' (2012) <https://d306pr3pise04h.cloudfront.net/docs/issues_doc%2Fhuman_rights%2FCRBP%2FChildrens_Rights_and_Business_Principles.pdf> accessed 30 November 2020, principle 6 (Use marketing and advertising that respect and support children's rights).

[177] See Pekka Mertala, 'How Connectivity Affects Otherwise Traditional Toys? A Functional Analysis of Hello Barbie' (2020) 25 Int. J. Child. Comput. Interact., DOI: 10.1016/j.ijcci.2020.100186 ("As Hello Barbie is able to speak, the child no longer performs the role through the doll, but in relation to the doll. This changes the nature of the performative element from dominantly transitive to dominantly performative, in which the child occupies and embodies a role in relation to the toy"). See also the following statement included in the list of all the lines Hello Barbie says as of November 17, 2015 (fn 158) "It's so cool that you want to be a mom someday".

[178] See 'Hello Barbie FAQ' (fn 159) ("The doll's conversation tree has been designed to re-direct inappropriate conversations. For example, Hello Barbie will not repeat curse words. Instead, she will respond by asking a new question"). However, besides the example given, there is no clear description of what is considered appropriate or not, and this category (appropriateness) is significantly influenced by the cultural component and potentially also by corporate ethics that may create forms of censorship or oriented behavior and thinking in the young user. Even when the FAQs refer to "school age appropriate content" ("All comments made by Hello Barbie are scripted with school age appropriate content"), they implicitly refer to a benchmark dependent the educational standards of developed economies.

assessment considering additional measures to mitigate unaddressed, or only partially addressed, concerns.

*6.1.2 Initial risk analysis and assessment*

The basic idea of the toy is an interactive doll, equipped with speech recognition and learning features, operating as an IoT device. The main component is a human-robot voice interaction feature based on AI and enabled by Internet connection and cloud services.

The rights potentially impacted are data protection and privacy, freedom of thought and diversity, and psychological and physical safety and health.[179]

*6.1.2.1 Data protection and the right to privacy*

While these are two distinct rights, for the purpose of this case study we considered them together.[180] Given the main product features, the impact analysis is based on following questions:[181]

- Does the device collect personal information? If yes, what kind of data is collected, and what are the main features of data processing? Can the data be shared with other entities/persons?
- Can the connected toy intrude into the users' private sphere?
- Can the connected toy be used for monitoring and surveillance purposes? If yes, is this monitoring continuous or can the user stop it?
- Do users belong to vulnerable categories (e.g. minors, elderly people, parents, etc.)?
- Are third parties involved in the data processing?
- Are transborder data flows part of the processing operations?

Taking into account the product's nature, features and settings (i.e. companion toy, dialogue recording, personal information collection, potential data sharing by parents) the likelihood of prejudice can be considered very high (Tab. 3). The extent and largely unsupervised nature of the dialogue between the doll and the user, as well as the extent of data collection and retention make the probability high (Tab. 1). In addition, given its default features and settings, the exposure is very high (Tab. 2) since all the doll's users are potentially exposed to this risk.

Regarding risk severity, the gravity of the prejudice (Tab. 4) is high, given the subjects involved (young children and minors), the processing of personal data in several main areas, including sensitive information,[182] and the extent of data collection. In addition, unexpected findings may emerge in the dialogue between the user and the doll, as the harmless topics prevalent in the AI-processed sentences can lead young users to provide personal and sensitive information.

---

[179] See Esther Keymolen and Simone Van der Hof, 'Can I still trust you, my dear doll? A philosophical and legal exploration of smart toys and trust' (2019) 4(2) Journal of Cyber Policy 143-159 ("Smart toys come in different forms but they have one thing in common. The development of these toys is not just a feature of ongoing technological developments; their emergence also reflects an increasing commercialisation of children's everyday lives").

[180] See also UN Convention on the Rights of the Child, Article 16; European Convention on Human Rights, Article 8.

[181] For a more extensive list of guiding questions, see e.g. UNICEF, 'Children's Online Privacy and Freedom of Expression' (2018) <https://www.unicef.org/csr/files/UNICEF_Childrens_Online_Privacy_and_Freedom_of_Expression(1).pdf> accessed 18 December 2020.

[182] Pre-recorded sentences containing references to, for instance, religion and ethical groups. See the full list of all lines for Hello Barbie (fn 158) (e.g. "Sorry, I didn't catch that. Was that a yes or a no to talking about Kwanzaa?").

Furthermore, the data processing also involves third parties and transborder data flows, which add other potential risks.

The effort to overcome potential prejudice or to reverse adverse effects (Tab. 5) can be considered as medium, due to the potential parental supervision and remote control, the nature of the doll's pre-selected answers and the adoption of standard data security measures that help to overcome suffered prejudice with a few difficulties (e.g. data erasure, dialogue with the minor in case of unexpected findings). Combining high gravity and medium effort, the resulting severity (Tab. 6) is medium.

If the likelihood of prejudice can be considered very high and the severity medium, the overall impact according to Tab. 8 is high.

*6.1.2.2 Freedom of thought, parental guidance and the best interest of the child*

Based on the main features of the product, the following questions can be used for this analysis:

- Is the device able to transmit content to the user?
- Which kind of relationships is the device able to create with the user?
- Does the device share any value-oriented messages with the user?
    - If yes, what kind of values are communicated?
    - Are these values customisable by users (including parents) or on the basis of user interaction? If so, what range of alterative value sets is provided?
    - Are these values the result of work by a design team characterised by diversity?

Here the case study reveals the critical impact of AI on HRI owing to the potential content imparted through the device. This is even more critical in the context of toys where the interactive nature of AI-powered dolls changes the traditional interaction into a relational experience.[183]

In the model considered (Hello Barbie), AI creates a dialogue with the young user by selecting the most appropriate sentence from the more than 8,000 lines of dialogue available in its database. On the one hand, this enables the AI to express opinions which may also include value-laden messages, as in this sentence: "It's so cool that you want to be a mom someday".[184] On the other, some value-based considerations are needed to address educational issues concerning "inappropriate questions"[185] where the problem is not the AI reaction (Hello Barbie responds "by asking a new question"[186]), as previously, but the notion of appropriateness, which necessarily involves a value-oriented content classification by the AI system.

As these value-laden features of AI are inevitably defined during the design process, the composition of the design team, its awareness of cultural diversity and pluralism are key elements that impact on freedom of thought, in terms of default values proposed and the availability of alternative settings. In addition, the decision to provide only one option or several user-customisable options in the case of value-oriented content is another aspect of the design phase that can limit parents' freedom to ensure the moral and religious education of their children in accordance with their own beliefs.

---

[183] See Mertala (fn 172).
[184] See fn 158. On gender stereotypes in smart toys, see Norwegian Consumer Council, *#Toyfail An analysis of consumer and privacy issues in three internet-connected toys* (2016) <https://fil.forbrukerradet.no/wp-content/uploads/2016/12/toyfail-report-desember2016.pdf> accessed 14 December 2020.
[185] See fn 174.
[186] See 'Hello Barbie FAQ' (fn 159).

This aspect highlights the paradigm shift brought by AI to freedom of thought and the related parental guidance in supporting the exercise by children of their rights.[187] This is even more evident when comparing AI-equipped toys with traditional educational products, such as books, serious games etc., whose contents can be examined in advance by parents.[188]

The AI-equipped doll is different. It delivers messages to young users, which may include educational content and information, but no parent will read all the 8,000 lines the doll can use or ask to have access to the logic used to match them with children's statements.

As AI-based devices interact autonomously with children and convey their own cultural values,[189] this impacts on the rights and duties of parents to provide, in a manner consistent with the evolving capacities of the child, appropriate direction and guidance in the child's freedom of thought, including aspects concerning cultural diversity.

In terms of risk assessment, the probability (Tab. 1) is medium, considering the limited number of sentences involving a value-oriented statement and the exposure (Tab. 2) is medium, due to their alignment with values commonly accepted in many cultural contexts. The likelihood is therefore medium (Tab. 3).

Taking into account the nature of the product and its main features (i.e. some value-laden sentences used in dialogue with the young user)[190] the gravity of prejudice (Tab. 4) can be considered low in the case in question, as the value-laden sentences concern cultural questions that are not particularly controversial. The effort (Tab. 5) can also be considered low, as talking with children can mitigate potential harm. Combining these two values, the severity is therefore low.

Note that this assessment would be completely altered if the dialogue content were not pre-selected, but generated by AI on the basis of information resulting from web searches,[191] where the potential risk would be much higher. Similarly, the inclusion in the pre-recorded database of a greater number of value-laden sentences would directly increase the risk.

Considering the likelihood as medium and the severity of the prejudice as low, the overall impact (Tab. 8) is medium.

*6.1.2.3 Right to psychological and physical safety*

Connected toys may raise concerns about a range of psychological and physical harms deriving from their use, including access to data and remote control of the toy.[192] Based on the main features of the product examined, the following questions can be used for this analysis:

---

[187] See UN Convention on the Rights of the Child, Articles 5, 14, and 18. See also See UNICEF, *Children's Online Privacy and Freedom of Expression* (2018), 9 <https://www.unicef.org/csr/files/UNICEF_Childrens_Online_Privacy_and_Freedom_of_Expression(1).pdf> accessed 18 December 2020, 9; Jim Murdoch, 'Protecting the Right to Freedom of Thought, Conscience and Religion under the European Convention on Human Rights' (Council of Europe 2012), 13.
[188] See also UN Convention on the Rights of the Child, Articles 17(e) and 18.
[189] See e.g. Norwegian Consumer Council (fn 179) referring to the connected dol Cayla ("Norwegian version of the apps has banned the Norwegian words for "homosexual", "bisexual", "lesbian", "atheism", and "LGBT" […]" "Other censored words include 'menstruation', 'scientology-member', 'violence', 'abortion', 'religion', and 'incest' ").
[190] See Valerie Steeves, 'A dialogic analysis of Hello Barbie's conversations with children' (2020) 7(1) Big Data & Society, DOI: 10.1177/2053951720919151.
[191] In the case examined, the content provided by means of the doll was handcrafted by the writing team at Mattel and ToyTalk, not derived from open web search. See 'Hello Barbie FAQ' (fn 159).
[192] See e.g. Otávio de Paula Albuquerque et al., 'Privacy in smart toys: Risks and proposed solutions' (2020) 39 Electronic Commerce Research and Applications, DOI: 10.1016/j.elerap.2019.100922, whose

- Can the device put psychological or physical safety at risk?
- Does the device have adequate data security and cybersecurity measures in place?
- Can third parties perpetrate malicious attacks that pose a risk to the psychological or physical safety of the user?

As regards the probability, considering the third-party origin of the prejudices and the limited interest in malicious attacks (no business interest, distributed and generic target), but also how easy it is to hack the toy, the probability (Tab. 1) of an adverse impact is medium. Exposure (Tab. 2) is low, given the prevalent use of the device in a supposedly safe environment, such as schools and home, where malicious access and control of the doll is difficult and adult monitoring is more frequent. The likelihood (Tab. 3) is therefore low.

Taking into account the nature of the product examined, the young age of the user, and the potential safety and security risks,[193] the gravity of prejudice (Tab. 4) can be considered medium. This is because malicious attacks can only be carried out by speech, and no images are collected. Nor can the toy – given its size and characteristics – directly cause physical harm to the user. The effort (Tab. 5) can be considered medium since parent-child dialogue and technical solutions can combat the potential prejudice. The severity (Tab. 6) is therefore medium.

Considering the likelihood as low and the severity of the prejudice as medium, the overall impact is medium (Tab. 8).

*6.1.2.4 Results of the initial assessment*

The following table shows the results of the assessment carried out on the initial idea of the connected AI-equipped doll described above:

Tab. 10. Table of envisaged risks for the examined case (L: low, M: medium; H: high; VH: very high)

| Risk | L | S | Overall impact |
|---|---|---|---|
| Impact on privacy and data protection | VH | M | H |
| Impact on freedom of thought | M | L | M |
| Impact on the right to psychological and physical safety | L | M | M |

Based on this table, we can plot a radial graph representing the overall impact on all the affected rights and freedoms. The graph below shows the priority of mitigating potentially adverse impacts on privacy and data protection, followed by risks related to physical integrity and freedom of thought.

This outcome is confirmed by the history of the actual product, where the biggest concerns of parents and the main reasons for its withdrawal related to personal data and hacking.[194]

---

authors refer to harassment, stalking, grooming, sexual abuse, exploitation, pedophilia and other types of violence blackmail, insults, confidence loss, trust loss and bullying; Shasha et al. (fn 161). See also Federal Bureau of Investigation, 'Consumer Notice: Internet-Connected Toys Could Present Privacy and Contact Concerns for Children' Alert Number I-071717 (Revised)-PSA (2017) <https://www.ic3.gov/Media/Y2017/PSA170717> accessed 15 December 2020.
[193] See fn 156.
[194] See Samuel Gibbs, *Hackers can hijack Wi-Fi Hello Barbie to spy on your children*, The Guardian, November 26, 2015 <https://www.theguardian.com/technology/2015/nov/26/hackers-can-hijack-wi-fi-hello-barbie-to-spy-on-your-children> accessed 12 November 2020.

Graph 2. Radial graph (impact) of the examined case

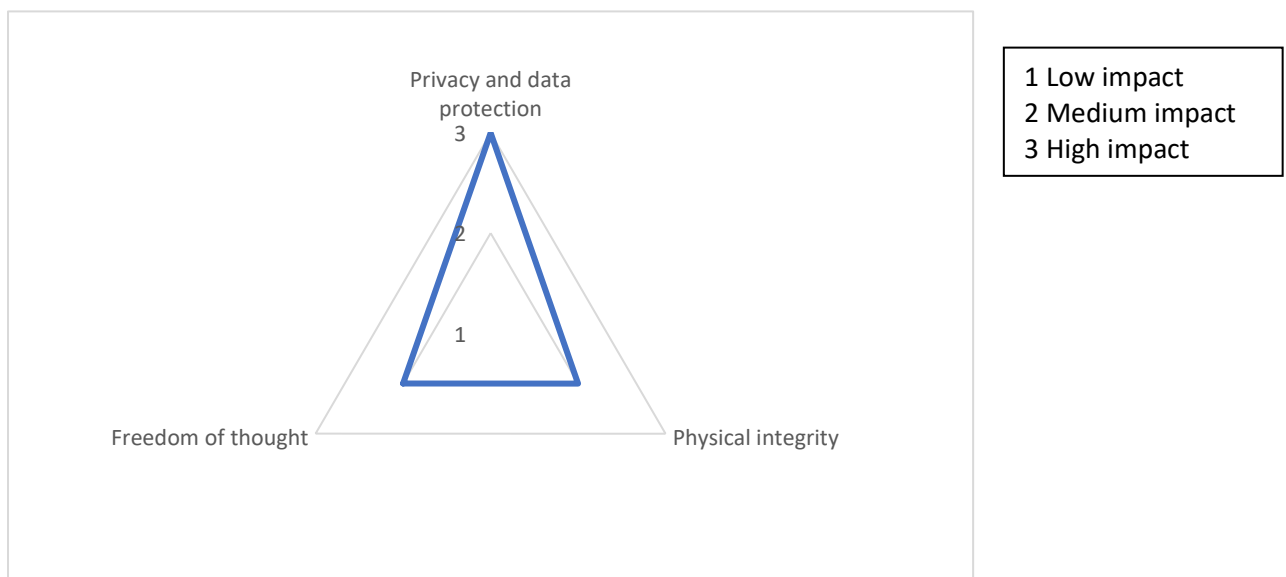

*6.1.3 Mitigation measures and re-assessment*

Following the iterative assessment, we can imagine that after this initial evaluation of the general idea, further measures are introduced to mitigate the potential risks found. At this stage, the potential stakeholders (users, parents associations, educational bodies, data protection authorities etc.) can make a valuable contribution to better defining the risks and how to tackle them.

While the role of the stakeholders cannot be directly assessed in this analysis, we can assume that their participation would have shown great concern for risks relating to communications privacy and security. This conclusion is supported by the available documentation on the reactions of parents and supervisory authorities.[195]

After the first assessment and given the evidence on stakeholders' requests, the following mitigation measures and by-design solutions could have been adopted with respect to the initial prototype.

A) Data protection and the right to privacy

Firstly, the product must comply with the data protection regulation of the countries in which it is distributed.[196] Given the product's design, we cannot exclude the processing of personal data.

---

[195] See e.g. BEUC, 'Connected Toys Do Not Meet Consumer Protection Standard. Letter to Mr Giovanni Buttarelli, European Data Protection Supervisor' (6 December 2016) <https://www.beuc.eu/publications/beuc-x-2016-136_mgo_letter_to_giovanni_buttarelli_-_edps_-_connected_toys.pdf> accessed 12 November 2020; ABA Journal, 'Moms Sue Mattel, Saying "Hello Barbie" Doll Violates Privacy' ABA Journal (9 December 2015) <https://www.abajournal.com/news/article/hello_barbie_violates_privacy_of_doll_owners_playmates_moms_say_in_lawsuit> accessed 20 March 2021; Emily McReynolds et al., 'Toys That Listen: A Study of Parents, Children, and Internet-Connected Toys', Proceedings of the 2017 CHI Conference on Human Factors in Computing Systems (ACM 2017) <https://dl.acm.org/doi/10.1145/3025453.3025735> accessed 12 November 2020.

[196] In this regard Hello Barbie was certified as compliant with the US COPPA, see 'Hello Barbie FAQ' (fn 159).

The limited number of sentences provided for use by AI, as in the case of Hello Barbie, does not exclude the provision of unexpected content by the user, including personal information.[197]

Risk mitigation should therefore focus on the topics of conversation between the doll and the young user, and the safeguards in processing information collected from the user.

As regards the first aspect, an effective way to limit the potential risks would be to use a closed set of sentences, excluding phrases and questions that might induce the user to disclose personal information, and making it possible to modify these phrases and questions by the owner of the toy.[198]

Regarding the processing of personal data, the doll's AI-based information processing functions should be deactivated by default, giving the parents control over its activation.[199] In addition, to reduce the risk of constant monitoring, deliberate action by the child should be required to activate the doll's AI-equipped dialogue functions.[200] This would also help to make users more aware of their interaction with the system and related privacy issues.[201]

Ex post remedies can also be adopted, such as speech detection to remove personal information in recorded data.[202]

Conversations are not monitored, except to support requests from parents. To reduce the impact on the right to privacy and data protection, human review of conversations – to test, improve, or change the technology used – should be avoided, even if specific policies for unexpected findings have been adopted.[203] Individual testing phases or experiments can be carried out in a laboratory setting or on the basis of user requests (e.g. unexpected reactions and dialogues). This more restrictive approach helps to reduce the impact with respect to the initial design.

Further issues, regarding the information processing architecture and its compliance with data protection principles, concern data storage. This should be minimised, and giving parents the possibility to delete stored information.[204]

---

[197] See 'Hello Barbie FAQ' (fn 159) ("we cannot control whether a child volunteers such information without prompting").

[198] In this case, the conditions are largely present, although there is evidence of minor issues. See e.g. 'Hello Barbie FAQ' (fn 159) ("Hello Barbie does not ask for a child's name and is not scripted to respond with a child's name, so she will not be able to recite a child's name back to them"), but see the interaction reported in Leta Jones (fn 162), 245 ("Barbie: Sometimes I get a little nervous when I tell people my middle name. But I'm really glad I told you! What's your middle name?! !"). The 'Hello Barbie FAQ' (fn 159) also points out the privacy-oriented design of the product with regard to dialogue content: "Although Hello Barbie was designed not to ask questions which are intended to elicit answers that might contain personal information".

[199] See 'Hello Barbie FAQ' (fn 159) ("Hello Barbie only requires a parent's email address to set up an account. This is necessary so that parents can give permission to activate the speech recognition technology in the doll. Other information, such as a daughter's birthday, can be provided to help personalize the experience but are not required […] If we discover that, in violation of our terms of service, an account was created by a child, we will terminate the account and delete all data and recordings associated with it.").

[200] In the Hello Barbie case, the doll was not always on but it was activated by pressing the belt buckle.

[201] In the examined case this was also emphasized because the two tri-color LEOs embedded in the doll's necklace lighted up to indicate she was active.

[202] See 'Hello Barbie FAQ' (fn 159) ("If we become aware of any such personal information captured in recordings, it is our policy to delete such information, and we contractually require our Service Providers to do the same. This personal information is not used for any purpose").

[203] See fn 163.

[204] See 'Hello Barbie FAQ' (fn 159) ("Parents who are concerned about this can monitor their child's use of Hello Barbie, and parents have the power to review and delete any conversation their child has with Hello Barbie, whether the conversations contain personal information or not"). Considering the young age

With regard to the use of collected data, while access to, and sharing of, this information by parents[205] are not per se against the interest of the child, caution should be exercised in using this information for marketing purposes. Given the early age of the users and the potentially large amount of information they may provide in their conversation with the doll, plus the lack of active and continuous parental control, the best solution would be not to use child-doll conversations for marketing.[206]

The complexity of data processing activities in the interaction between a child and an AI-equipped doll inevitably affects the form and content of the privacy policies and the options offered to users, as provided by many existing legislations.

A suitable notice and consent mechanism, clear and accessible and legally compliant, is therefore required,[207] but meeting this obligation is not so simple in the case in question. The nature of the connected toy and the absence of any interface limits awareness of the policies and distances them from direct interaction with the device. This accentuates the perception of the notice and consent mechanism as a mere formality to be completed to access the product.

The last crucial area concerns data security. This entails a negative impact that goes beyond personal data protection and, as such, is also analysed below under impact on the right to psychological and physical safety.

As the AI-based services are hosted by the service provider, data security issues concern both device-service communications and malicious attacks to the server and the device. Encrypted communications, secure communication solutions, and system security requirements for data hosted and processed on the server can minimise potential risks, as in the case study, which also considered access to data when the doll's user changes.[208]

None of these measures prevent the risks of hacking to the device or the local Wi-Fi connection, which are higher when the doll is used outdoors.[209] This was the chief weakness noted in the case in question and in IoT devices more generally. They are often designed with poor inherent data security and cybersecure features for cost reasons. To reduce this risk, stronger authentication and encryption solutions have been proposed in the literature.[210]

Taking into account the initial impact assessment plus all the measures described above, the exposure is reduced to low, since users are thus exposed to potential prejudices only in special

---

of the user this seems not to be a disproportionate monitoring with regard to their activities and right to privacy. This does not exclude a socio-ethical relevance of this behavior, see e.g. Meg Leta Jones and Kevin Meurer, 'Can (and Should) Hello Barbie Keep a Secret?' (2016) IEEE International Symposium on Ethics in Engineering, Science and Technology (ETHICS), doi: 10.1109/ETHICS.2016.7560047 ("the passive nature of Barbie's recording capabilities could prove perhaps more devastating to a child who may have placed an implicit trust in the doll. In order to determine the extent of the parent's involvement in their child's recordings, we extended our analysis to include the adult oversight capabilities").
[205] See above fn 166.
[206] This was the option adopted in the Hello Barbie case, see fn. 167. But see Steeves (fn 185) on the sentences used by Hello Barbie to indirectly reinforce the brand identity and encourage the child to adopt that identity for his/her own.
[207] In the case examined, one of the main weakness claimed with regard to Hello Barbie concerned the privacy policies adopted, the interplay between the different entities involved in data processing, and the design of these policies and access to them, which were considered cumbersome. See Leta Jones and Meurer (fn 199).
[208] See also 'Hello Barbie FAQ' (fn 159) ("Conversations and other information are not stored on the doll itself, but rather in the associated parent account. So, if other users are using a different WiFi network and using their own account, Hello Barbie would not remember anything from the prior conversations. New users would need to set up their own account to enable conversations with Barbie").
[209] See Leta Jones (fn 162), 244.
[210] See also below letter C).

circumstances, primarily malicious attack. Probability also becomes low, as the proposed measures mitigate the risks relating to dialogue between doll and user, data collection and retention. Likelihood (Tab. 3) is therefore reduced to low.

Regarding severity of prejudice, gravity can be lowered to at least medium by effect of the mitigation measures, but effort remains medium, given the potential risk of hacking. Severity is therefore lowered somewhat (from 5 to 3 in Tab. 6), though remaining medium.

If the severity and the likelihood are medium in Tab. 8, the overall impact is lowered from high to medium.

B) Impact on freedom of thought

As described in Section 6.1.2.2, the impact on freedom of thought is related to the values conveyed by the doll in dialogue with the user. Here the main issue concerns the nature of the messages addressed to the user, their sources and their interplay with the rights and duties of parents to provide appropriate direction and guidance in the child's exercise of freedom of thought, including issues of cultural diversity.

A system based on Natural Language Processing allows AI various degrees of autonomy in identifying the best response or sentence in the human-machine interaction. Given the issues considered here (the nature of the values shared by the doll with its young user) the two main options are to use a closed set of possible sentences or search for potential answers in a large database, such as the Internet. A variety of solutions can also be found between these two extremes.

Since the main problem is content control, the preferable option is the first, and this was indeed the solution adopted in the Hello Barbie case.[211] Content can thus be fine-tuned to the education level of the user, given the age range of the children.[212] This reduces the risk of unexpected and inadequate content and, where full lines of dialogue are available (this was the case with Hello Barbie), parents are able to get an idea of the content offered to their children.

Some residual risks remain however, due to intentional or unintentional cultural models or values, including the difference between appropriate and inappropriate content.[213] This is due to the special relationship the toy generates[214] and the only limited mitigation provided by transparency on pre-recorded lines of dialogue.

To address these issues, concerning both freedom of thought and diversity, the AI system should embed a certain degree of flexibility (user-customizable content) and avoid stereotyping by default. To achieve this, the team working on pre-recorded sentences and dialogues should be characterised by diversity, adopting a by-design approach and bearing in mind the target user of the product.[215]

Moreover, taking into account the parents' point of view, mere transparency, i.e. access to the whole body of sentences used by the doll, is not enough. As is demonstrated extensively in the field of data protection, information on processing is often disregarded by the user and it is hard to imagine parents reading 8,000 lines of dialogue before buying a doll.

---

[211] 'Hello Barbie FAQ' <http://hellobarbiefaq.mattel.com/faq/> accessed 12 November 2020.
[212] See also 'Hello Barbie FAQ' (fn 159) ("All comments made by Hello Barbie are scripted with school age appropriate content").
[213] See fn 173.
[214] See fn 172.
[215] See Steeves (fn 185) on the different attitude in prerecorded sentences with regard to different religious topics.

To increase transparency and user awareness, therefore, forms of visualisation of these values through logic and content maps could be useful to easily represent the content used. In addition, it would be important to give parents the opportunity to partially shape the AI reactions, customising the values and content, providing other options relating to the most critical areas in terms of education and freedom of thought.

With regard to the effects of these measures, they mitigate both the potentially adverse consequences of product design and the lack of parental supervision of content, minimising the probability of an adverse impact on freedom of thought. The probability (Tab. 1) is therefore lowered to low.

Given the wide distribution of the product, the potential variety of cultural contexts and the need for an active role of parents to minimise the risk, the exposure remains medium, although the number of affected individuals is expected to decrease (Tab. 2).

If the probability is low and the exposure is medium, the likelihood (Tab. 3) is lowered to low after the adoption of the suggested mitigation measures and design solutions.

The gravity of prejudice and the effort were originally low and the additional measures described can further reduce gravity through a more responsible management of content which might support potentially conflicting cultural models or values. Severity therefore remains low.

Considering both likelihood and severity as low, the overall impact (Tab. 8) is reduced from medium to low, compared with the original design model.

C) Impact on the right to psychological and physical safety

The potential impact in this area is mainly related to malicious hacking activities[216] that might allow third parties to take control of the doll and use it to cause, psychological and physical harm to the user.[217] This was one of the most widely debated issues in the Hello Barbie case and one of the main reasons that led Mattel to stop producing this toy.[218] Possible mitigation measures are the exclusion of interaction with other IoT devices,[219] strong authentication and data encryption.[220]

As regards likelihood, considering the protection measures adopted and the low interest of third parties in this type of individual and context-specific malicious attack, the probability is low (Tab. 1). Although the suggested measures do not affect the exposure, this remains low due to the limited circumstances in which a malicious attack can be carried out (Tab. 2). The likelihood therefore remains low but is lowered (from 2 to 1 in Tab. 3).

---

[216] See Gibbs (fn 189).
[217] See Victor Chang, Zhongying Li and Muthu Ramachandran, 'A Review on Ethical Issues for Smart Connected Toys in the Context of Big Data' in Farshad Firouzi, Ernesto Estrada, Victor Mendez Munoz, Victor Chang, *COMPLEXIS 2019 - Proceedings of the 4th International Conference on Complexity, Future Information Systems and Risk* (SciTePress 2019) 149-156 ("For example, the attackers can spread content through the audio system, which is adverse for children's growth through the built-in audio in the smart toys").
[218] See also Shasha et al. (fn 161).
[219] Doll's speech content was hand crafted by the writing team at Mattel and ToyTalk, not derived by open web search. See 'Hello Barbie FAQ' (fn 159).
[220] See Katerina Demetzou, Leon Böck and Obaida Hanteer, 'Smart Bears don't talk to strangers: analysing privacy concerns and technical solutions in smart toys for children' in *IET Conference Proceedings; Stevenage Stevenage: The Institution of Engineering & Technology* (2018) DOI:10.1049/cp.2018.0005; Luciano Gonçalves de Carvalho and Marcelo Medeiros Eler, 'Security Tests for Smart Toys' in Proceedings of the 20th International Conference on Enterprise Information Systems (2018) 111-120 <http://www.scitepress.org/DigitalLibrary/Link.aspx?doi=10.5220/0006776101110120> accessed 23 December 2020.

Regarding severity, the proposed measures do not impact on the gravity of the prejudice (Tab. 4), or the effort (Tab. 5) which remain medium. Severity therefore remains medium (Tab. 6).

Since the final values of neither likelihood nor severity change, overall impact remains medium (Tab. 8), with malicious hacking being the most critical aspect of the product in terms of risk mitigation.

The following table shows the assessment of the different impacts, comparing the results before and after the adoption of mitigation measures.

Tab. 11. Comparative risk impact analysis table (examined case)

| Risk | L | S | Overall impact | MMs | rL | rS | Final impact |
|---|---|---|---|---|---|---|---|
| Impact on privacy and data protection | VH | M | H | See above sub A) | M | M | M |
| Impact on freedom of thought | M | L | M | See above sub B) | L | L | L |
| Impact on the right to psychological and physical safety | L | M | M | See above sub C) | L | M | M |
| Overall impact (all impacted areas) | | | M/H | | | | M/L |

In the case in question, there is no Tab. 9 EF column since there are no factors that could exclude risk, such as certain mandatory impacting features or overriding competing interests recognised by law.

Graph 3. Final radial graph of the examined case

[Blue line: original impact. Orange line: final impact after adoption of mitigation measures and design solutions]

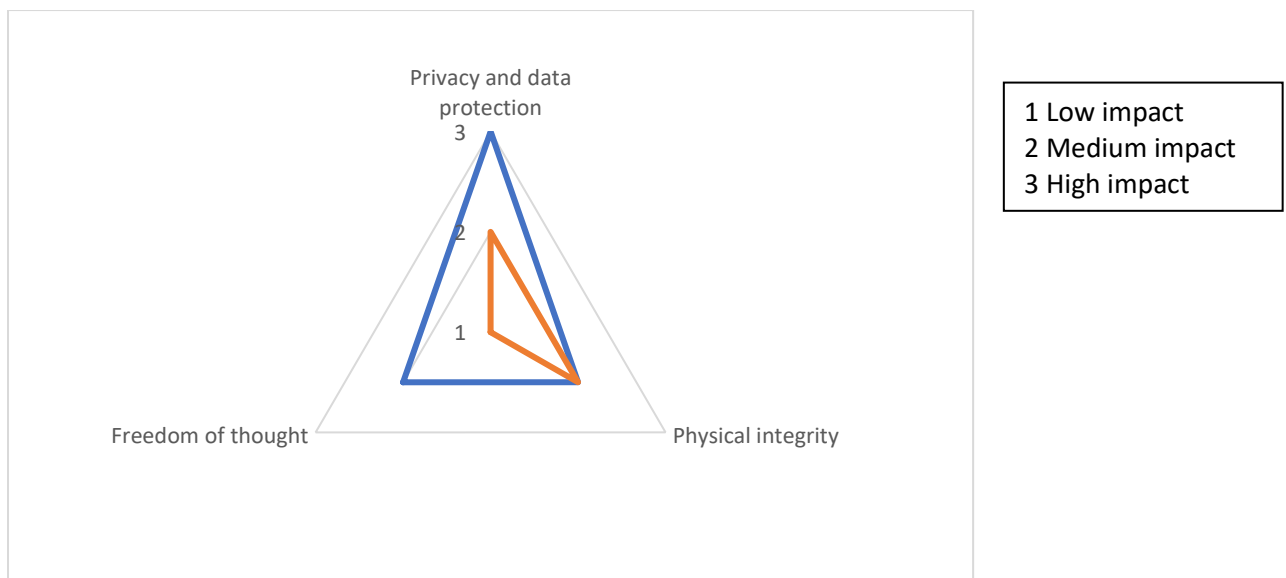

This graph shows the concrete effect of the assessment (the blue line represents the initial impacts and the orange the impacts after adoption of the measures described above). It should be noted that the reduction of potential impact is limited as the Hello Barbie product already included several options and measures to mitigate adverse effects on rights and freedoms (pre-recorded sentences, no Internet access, data encryption, parental access to stored data, etc.). The effect would have been greater starting from a general AI-equipped doll using Natural Language Processing interacting with children, without mitigation measures.

In this regard, the HRIA model we propose is in line with a human rights-by design approach, where the design team is asked to consider human rights impact from the earliest product design stages, discarding those options that have an obvious negative impact on human rights. With this approach, there is no HRIA 0 where the proposed product is completely open to the riskiest scenarios (e.g. a connected doll equipped with unsupervised AI that uses all available web sources to dialogue with young users, with unencrypted doll-user communication sent to a central datacentre where information is stored without a time limit and used for further purposes, including marketing communications direct to doll users).

In human rights-oriented design, HRIA thus becomes a tool to test, refine and improve adopted options that already entail a risk-aware approach. In this way, HRIA is a tool for testing and improving human rights-oriented design strategies.

*6.2 HRIA in large-scale multi-factor scenarios: the Sidewalk case*

Large-scale projects using data-intensive applications are characterised by a variety of potentially impacted areas concerning individual and groups. This produces a more complex and multi-factor scenario which cannot be fully assessed by the mere aggregation of the results of HRIAs conducted for each component of data-intensive applications.

An example is provided by data-driven smart cites, where the overall effect of an integrated model including different layers affecting a variety of human activities means that the cumulative impact is greater than the sum of the impacts of each application.

In such cases, a HRIA for data-intensive systems also needs to consider the cumulative effect of data use and the AI strategies adopted, as already happens in HRIA practice with large-scale scenario cases. This is all the more important in the field of AI where large-scale projects often feature a unique or dominant technology partner who benefits from a general overview of all the different processing activities ('platformisation'[221]).

The Sidewalk project in Toronto is an example of this 'platformisation' effect and a case study in the consequent impacts on rights and freedoms. This concluded smart city project was widely debated[222] and raised several human rights-related issues common to other data-intensive projects. It also highlights how the universal nature of the benchmark framework proposed makes the assessment model suited to deployment in various jurisdictions, beyond European borders.

The case concerned a requalification project for the Quayside, a large urban area on Toronto's waterfront largely owned by Toronto Waterfront Revitalization Corporation. Based on an agreement between the City of Toronto and Toronto Waterfront,[223] in 2017, through a competitive

---

[221] See Ellen Goodman & Julia Powles, *Urbanism Under Google: Lessons from Sidewalk Toronto*, 88 Fordham Law Review 457–498 (2019).
[222] See Constance Carr and Markus Hesse, 'When Alphabet Inc. Plans Toronto's Waterfront: New Post-Political Modes of Urban Governance' (2020) 5 Urban Planning 69-83; Alexandra Flynn and Mariana Valverde, 'Where The Sidewalk Ends: The Governance Of Waterfront Toronto's Sidewalk Labs Deal' (2019) 36 Windsor Yearbook of Access to Justice 263–283.
[223] The Waterfront Revitalization Corporation (which was renamed Waterfront Toronto) was a partnered not-for-profit corporation, created in 2003 by the City of Toronto, Province of Ontario and the Government of Canada (see also Province's Toronto Waterfront Revitalization Corporation Act) to oversee and deliver revitalization of Toronto's waterfront; further information are available here: <https://www.toronto.ca/city-government/accountability-operations-customer-service/city-

Request for Proposals, Waterfront Toronto hired Sidewalk Labs (a subsidiary of Alphabet Inc.) to develop a proposal for this area.[224]

This proposal – the Master Innovation and Development Plan or MIDP[225] – outlined a vision for the Quayside site and suggested data-driven innovative solutions across the following areas: mobility and transportation; building forms and construction techniques; core infrastructure development and operations; social service delivery; environmental efficiency and carbon neutrality; climate mitigation strategies; optimisation of open space; data-driven decision making; governance and citizen participation; and regulatory and policy innovation.[226]

This long list of topics shows how the data-intensive project went beyond mere urban requalification to embrace goals that are part of the traditional duties of a local administration, pursuing public interest purposes[227] with potential impacts on a variety of rights and freedoms.

The Sidewalk case[228] suggests several takeaways for the HRIA model. First, an integrated model, which combines the HRIAs of the different technologies and processes adopted within a multi-factor scenario, is essential to properly address the overall impact, including a variety of socio-technical solutions and impacted areas.

---

administration/city-managers-office/agencies-corporations/corporations/waterfront-toronto/> accessed 30 December 2020. See also Toronto Waterfront Revitalization: Memorandum of Understanding between the City of Toronto, City of Toronto Economic Development Corporation and Toronto Waterfront Revitalization Corporation
<https://www.toronto.ca/legdocs/2006/agendas/council/cc060131/pof1rpt/cl027.pdf> accessed 30 December 2020; City of Toronto, Executive Committee, 'Executive Committee consideration on January 24' 2018.EX30. 9 (2018) <http://app.toronto.ca/tmmis/viewAgendaItemHistory.do?item=2018.EX30.9> (last visited Dec 30, 2020).
[224] Waterfront Toronto and Sidewalk Labs entered into a partnership Framework Agreement on October 16, 2017. The Framework Agreement was a confidential legal document, see City of Toronto, Executive Committee (fn 223). A summary of this agreement is available in City of Toronto, Executive Committee, 'Executive Committee consideration on January 24, 2018, 2018.EX30. 9. Report and Attachments 1 and 2 from the Deputy City Manager, Cluster B on Sidewalk Toronto' (2018)
<https://www.toronto.ca/legdocs/mmis/2018/ex/bgrd/backgroundfile-110745.pdf> accessed 31 December 2020, Comments, para 2 and Attachment 2.
[225] Sidewalk Labs was charged with providing Waterfront Toronto with a MIDP for evaluation, including public and stakeholder consultation. Following the adoption of the MIDP by the Waterfront Toronto's Board of Directors, the City of Toronto was to complete an additional assessment programme focused on feasibility and legal compliance, including public consultation. See City of Toronto, Deputy City Manager, Infrastructure and Development, 'Report for action. EX6.1' (2019)
<https://www.toronto.ca/legdocs/mmis/2019/ex/bgrd/backgroundfile-133867.pdf> accessed 30 December 2020.
[226] See City of Toronto, Executive Committee, 'Executive Committee consideration on January 24, 2018, 2018.EX30. 9' (fn 224).
[227] See also Bianca Wylie, 'In Toronto, Google's Attempt to Privatize Government Fails—For Now', Boston Review, 13 May 2020; Goodman & Powles (fn 221).
[228] For a more extensive discussion of this case, see also Teresa Scassa, 'Designing Data Governance for Data Sharing: Lessons from Sidewalk Toronto' (2020) Special Issue: Governing Data as a Resource, Technology and Regulation 44-56; Kevin Morgan and Brian Webb, 'Googling the City: In Search of the Public Interest on Toronto's 'Smart' Waterfront' (2020) 5 Urban Planning 84–95; Anna Artyushina, 'Is civic data governance the key to democratic smart cities? The role of the urban data trust in Sidewalk Toronto' (2020) 55 Telematics and Informatics, DOI: 10.1016/j.tele.2020.101456; Flynn and Valverde (fn 222); Kyle Peel and Eliot Tretter, 'Waterfront Toronto: Privacy or Piracy?' (2019)
<https://osf.io/xgzs2s> accessed 28 December 2020; Constance Carr and Markus Hesse, 'Sidewalk Labs closed down – whither Google's smart city' (2020) Regions
<https://regions.regionalstudies.org/ezine/article/sidewalk-labs-closed-down-whither-googles-smart-city/> accessed 28 December 2020); Goodman & Powles (fn 221).

Second, the criticism surrounding civic participation in the Sidewalk project reveals how the effective engagement of relevant stakeholders is central from the earliest stages of proposal design. Giving voice to potentially affected groups mitigates the risk of the development of top-down and merely technology driven solutions, which have a higher risk of rejection and negative impact.

Third, the complexity and extent of large-scale integrated HRIA for multi-factor scenarios require a methodological approach that cannot be limited to an internal self-assessment but demand an independent third-party assessment by a multidisciplinary team of experts, as in traditional HRIA practice.

These elements suggest three key principles for large-scale HRIA: independence, transparency, and inclusivity. Independence requires third-party assessors with no legal or material relationship with the entities involved in the projects, including any potential stakeholders.

Transparency concerns both the assessment procedure, facilitating stakeholders' participation, and the public availability of the assessment outcome,[229] using easily understandable language. In this sense, transparency is linked to inclusivity, which concerns the engagement of all the different stakeholders impacted by the activities examined.

Tab. 12 Multi-factor scenario HRIA: main stages and tasks

| Main stage | Sub-section | Main tasks |
| --- | --- | --- |
| I. Planning and scoping | A. Preliminary analysis | 1. Collection of information on the project, parties involved (including supply-chain), potential stakeholders, and territorial target area (country, region).[230]<br>2. Human rights reference framework: review of applicable binding and non-binding instruments, gap analysis. |
|  | B. Scoping | 1. Identification of main issues related to human rights to be examined.<br>2. Drafting of a questionnaire for HRIA interviews and main indicators. |
| II. Risk analysis and assessment | A. Fieldwork | 1. Interviews with internal and external project stakeholders[231] and data collection.[232]<br>2. Understanding of contextual issues (political, economic, regulatory, and social). |

---

[229] See also Mantelero (fn 35) 766, fn 94("It is possible to provide business-sensitive information in a separate annex to the impact assessment report, which is not publicly available, or publish a short version of the report without the sensitive content").

[230] See also The Danish Institute for Human Rights, *Guidance on HRIA of Digital Activities. Phase 2: Data Collection and context analysis* (2020), 13-18 <https://www.humanrights.dk/sites/humanrights.dk/files/media/document/Phase%202_Data%20Collection%20and%20Context%20Analysis_n.pdf> accessed 20 February 2021.

[231] Various interview techniques can be used in the assessment, such as focus groups, women-only group interviews, one-on-one interviews (key persons) and interviews with external stakeholders.

[232] Taking into account the circumstances, e.g. vulnerable groups, data could be collected anonymously through written submissions.

|  | B. Analysis and assessment | 1. Data verification and validation, comparing and combining fieldwork results and desk analysis.<br>2. Further interviews and analysis, if necessary.<br>3. Impact analysis for each project branch and impacted rights and freedoms.<br>4. Integrated impact assessment report.[233] |
|---|---|---|
| III. Mitigation and further implementation | A. Mitigation | 1. Recommendations.<br>2. Prioritisation of mitigation goals. |
|  | B. Further implementation | 1. Post-assessment monitoring.<br>2. Grievance mechanisms.<br>3. Ongoing stakeholder engagement. |

An additional important contribution of the integrated HRIA is its ability to shed light on issues that do not emerge in assessing single components of large-scale AI systems, as the cumulative effect of such projects is key. Here, the human rights layer opens up to a broader perspective which includes the impact of socio-technical solutions on democratic participation and decisions.

The Urban Data Trust created by Sidewalk and its role in the Toronto project is an example in this sense. The Urban Data Trust was tasked with establishing "a set of RDU [Responsible Data Use] Guidelines that would apply to all entities seeking to collect or use urban data" and with implementing and managing "a four-step process for approving the responsible collection and use of urban data" and any entity that wishes to collect or use urban data in the district "would have to comply with UDT [Urban Data Trust] requirements, in addition to applicable Canadian privacy laws".[234]

This important oversight body was to be created by an agreement between Waterfront Toronto and Sidewalk Lab[235] and composed of a board of five members (a data governance, privacy, or intellectual property expert; a community representative; a public-sector representative; an academic representative; and a Canadian business industry representative) acting as a sort of internal review board and supported by a Chief Data Officer who, under the direction of the board, was to carry out crucial activities concerning data use.[236] In addition, the Urban Data Trust would have to enter into contracts with all entities authorised to collect or use urban data[237] in the district,

---
[233] See also The Danish Institute for Human Rights, '*Guidance on HRIA of Digital Activities. Phase 5: Reporting and Evaluation_n.pdf*' (2020) <https://www.humanrights.dk/sites/humanrights.dk/files/media/document/Phase%205_Reporting%20and%20Evaluation_n.pdf> accessed 20 February 2021.
[234] See Side Walk Labs, 'Toronto Tomorrow. A new approach for inclusive growth. MIDP' (2019) Vol. 2, 419 and Vol. 3, 69. On the interplay the role of the Urban Data Trust in setting requirements for data processing and the legal framework into force in Canada and in Toronto, see Scassa (fn 228).
[235] See also Sacassa (fn 228) 55 ("in proposing the UDT, Sidewalk Labs chose a governance model developed unilaterally, and not as part of a collective process involving data stakeholders").
[236] See Side Walk Labs (fn 234) vol. 2, 421 ("the Chief Data Officer would be responsible for developing the charter for the Urban Data Trust; promulgating RDU Guidelines that apply to all parties proposing to collect urban data, and that respect existing privacy laws and guidelines but also seek to apply additional guidelines for addressing the unique aspects of urban data […]; structuring oversight and review processes; determining how the entity would be staffed, operated, and funded; developing initial agreements that would govern the use and sharing of urban data; and coordinating with privacy regulators and other key stakeholders, as necessary").
[237] The notion of urban data is a novel category proposed by Sidewalk, referring to "both personal information and information that is not connected to a particular individual […] it is collected in a

and these data sharing agreements could also "potentially provide the entity with the right to enter onto property and remove sensors and other recording devices if breaches are identified".[238]

Although this model was later abandoned, due to the concerns raised by this solution,[239] it shows the intention to create an additional layer of data governance, different from both the individual dimension of information self-determination and the collective dimension of public interest managed by public bodies, within a process of centralisation and privatisation of data governance regarding information generated within a community.[240]

In this sense, the overall impact of AI applications in urban spaces and their coordination by a dominant player providing technological infrastructure raise important questions about the cumulative effect on potentially impacted rights, and even more concerning democracy and the socio-political dimension of the urban landscape,[241] particularly in terms of the division of public and private responsibilities on matters of collective interest.

This privatisation of the democratic decision process, based on the 'platformisation' of the city, directly concerns the use of data, but is no longer just about data protection. In socio-technical contexts, data governance is about human rights in general, insofar as the use of data by different AI applications raises issues about a variety of potentially adverse effects on different rights and freedoms.[242] If data becomes a means of managing and governing society, its use necessarily has an impact on all the rights and freedoms of individuals and society. This impact is further exacerbated by the empowerment enabled by AI technologies (e.g. the use of facial recognition to replace traditional video-surveillance tools).

For these reasons, cumulative management of different data-intensive systems impacting on the social environment cannot be left to private service providers or an ad hoc associative structure, but should remain within the context of public law, centred on democratic participation in decision-making processes affecting general and public interest.[243]

Large-scale data-intensive projects therefore suggest using the HRIA not only to assess the overall impact of all the various AI applications used, but also to go beyond the safeguarding of human rights and freedoms. The results of this assessment therefore become a starting point for a broader

---

physical space in the city and may be associated with practical challenges in obtaining meaningful consent […] Urban data would be broader than the definition of personal information and include personal, non-personal, aggregate, or de-identified data […] collected and used in physical or community spaces where meaningful consent prior to collection and use is hard, if not impossible, to obtain", see Side Walk Labs (fn 234) vol. 2, 416. But see, for critical comments on this category and its use, Scassa (fn 228) 51-54; Goodman & Powles (fn 221), 473.

[238] See Side Walk Labs (fn 234) vol. 2, 420-422.
[239] See Open Letter from Waterfront Toronto Board Chair, 31 October 2019 <https://waterfrontoronto.ca/nbe/wcm/connect/waterfront/waterfront_content_library/waterfront+home/news+room/news+archive/news/2019/october/open+letter+from+waterfront+toronto+board+chair+-+october+31%2C+2019> accessed 8 March 2021.
[240] See also Artyushina (fn 228).
[241] See also Carr and Hesse (fn 228).
[242] See e.g Raso et al. (fn 39).
[243] The right to participate in public affairs (Article 25 Covenant) is based on a broad concept of public affairs, which includes public debate and dialogue between citizens and their representatives, with close links to freedom of expression, assembly and association. See UN Human Rights Committee (HRC), CCPR General Comment No. 25: The right to participate in public affairs, voting rights and the right of equal access to *public* service (Art. 25), CCPR/C/21/Rev.1/Add.7, 12 July 1996.See also UN Committee on Economic, Social and Cultural Rights (CESCR), General Comment No. 1: Reporting by States Parties, 27 July 1981, para 5.

analysis and planning of democratic participation in the decision-making process on the use of AI, including democratic oversight on its application.[244]

In line with the approach adopted by international human rights organisations, the human rights dimension should combine with the democratic dimension and the rule of law in guiding the development and deployment of AI projects from their earliest stages.

The findings of the HRIA will therefore also contribute to addressing the so-called 'Question Zero' about the desirability of using AI solutions in socio-technical systems. This concerns democratic participation and the freedom of individuals, which are even more important in the case of technological solutions in an urban context, where people often have no real opportunity to opt out due to the solutions being deeply embedded in the structure of the city and its essential services.

A key issue then for the democratic use of AI concerns architecture design and its impact on rights and freedoms. The active role of technology in co-shaping human experiences[245] necessarily leads us to focus on the values underpinning the technological infrastructure and how these values are transposed into society through technology.[246] The technology infrastructure cannot be viewed as neutral, but as the result of both the values, intentionally or unintentionally, embedded in the devices/services and the role of mediation played by the different technologies and their applications.[247]

These considerations on the power of designers – which are widely discussed in the debate on technology design[248] – are accentuated in the context of smart cities and in many large-scale data-intensive systems. Here, the key role of service providers and the 'platformisation' of these environments[249] shed light on the part these providers play with respect to the overall impact of the AI systems they manage.

The HRIA can play an important role in assessing values and supporting a human rights-oriented design that also pays attention to participatory processes and democratic deliberation governing large-scale AI systems. This can facilitate the concrete development of a truly trustworthy AI, in which trust is based on respect for human rights, democracy and the rule of law.

---

[244] See also Alessandro Mantelero, 'Analysis of international legally binding instruments' in Council of Europe (fn 10), 82-88.
[245] See also Noëmi Manders-Huits and Jeroen van den Hoven, 'The Need for a Value-Sensitive Design of Communication Infrastructures' in Paul Sollie and Marcus Düwell (eds) *Evaluating New Technologies. Methodological Problems for the Ethical Assessment of Technology Developments* (Springer 2009), 55-56.
[246] See also Don Ihde, *Technology and the Lifeworld: from garden to earth* (Indiana University Press 1990).
[247] See Bruno Latour and Couze Venn, 'Morality and Technology: The End of the Means' (2002) 19(5-6) Theory, Culture and Society 247-60.
[248] See also Langdon Winner, 'Do Artifacts Have Politics?' (1980) Daedalus, 109, 121–136; Langdon Winner, 'Technē and Politeia: The Technical Constitution of Society' in Paul T. Durbin and Friedrich Rapp (eds) *Philosophy and Technology* (Springer 1983) 105 ("let us recognize that every technology of significance to us implies a set of political commitments that one can identify if one looks carefully enough. To state it more directly, what appear to be merely instrumental choices are better seen as choices about the form of the society we continually build, choices about the kinds of people we want to be"). See Verbeek (fn 7), 109, 129, and 164-165 ("Accompanying technological developments requires engagement with designers and users, identifying points of application fir moral reflection, and anticipating the social impact of technologies-in-design […] In order to develop responsible forms of use and design, we need to equip users ad designer with frameworks and methods to anticipate, assess, ad design the mediating role of technologies in people's lives and in the ways we organize society").
[249] See Council of Europe, Consultative Committee of the Convention 108 (T-PD), 'Guidelines on Artificial Intelligence and Data Protection', Strasbourg, 25 January 2019, T-PD(2019)01; Council of Europe, Committee of Ministers (fn 39).

# 7. Conclusions

The recent turn in the debate on AI regulation from ethics to law, the wide application of AI and the new challenges it poses in a variety of fields of human activities are urging legislators to find a paradigm of reference to assess the impacts of AI and to guide its development. This cannot only be done at a general level, on the basis of guiding principles and provisions, but the paradigm must be embedded into the development and deployment of each application.

With a view to providing a global approach in this field, human rights and fundamental freedoms can offer this reference paradigm for a truly human-centred AI. However, this growing interest in a human rights-focused approach needs to be turned into effective tools that can guide AI developers and key AI users, such as municipalities, governments, and private companies.

To bridge this gap between theoretical thinking on the potential role of human rights in addressing and mitigating AI-related risks, this work has suggested an empirical evidence-based approach to developing a human rights impact assessment (HRIA) model for AI.

Using the results of an in-depth analysis of jurisprudence in the field of data processing in Europe, we have outlined how human rights and freedoms already play an important role in the assessment of data-intensive applications. However, there is the lack of a formal methodology to facilitate an ex-ante approach based on a human-oriented design of product/service development. Moreover, this empirical analysis has better clarified the interplay between human rights and data processing in data-intensive systems, facilitating the development of an evidence-based model that is easier to implement as it is based on existing case law rather than on an abstract theoretical evaluation of the potential impact of AI.

The core of our research is the proposed HRIA model for AI, which has been developed in line with the existing practices in HRIA, but in a way that better responds to the specific nature of AI applications, in terms of scale, impacted rights and freedoms, prior assessment of production design, and assessment of risk levels, as required by several proposals on AI regulation.

The result is a tool that can be easily used by entities involved in AI development from the outset in the design of new AI solutions and can follow the product/service throughout its lifecycle, providing specific, measurable and comparable evidence on potential impacts, their probability, extension, and severity, and facilitating comparison between alternative design options and an iterative approach to AI design, based on risk assessment and mitigation.

In this sense, the proposed model is no longer just an assessment tool but a human rights management tool, providing clear evidence for a human rights-oriented development of AI products and services and their risk management.

In addition, a more transparent and easy-to-understand impact assessment model facilitates a participatory approach to AI development by potential stakeholders, giving them clear and structured information about possible options and the effects of changes in AI design.

Finally, the proposed model can also be used by supervisory authorities and auditing bodies to monitor risk management in relation to the impact of data use on individual rights and freedoms.

Based on these results, several concluding remarks can be drawn. The first general one is that conducting a HRIA should be seen not as a burden or a mere obligation, but as an opportunity. Given the nature of AI products/services and their features and scale, the proposed assessment model can significantly help companies and other entities to develop effective human-centric AI in challenging contexts.

The model can also contribute to a more formal and standardised assessment of AI solutions, facilitating comparison between different options and design approaches. Although HRIA has already been adopted in several contexts, large-scale projects are often assessed without using a formal evaluation of risk likelihood and severity.[250] Traditional HRIA reports often describe the risks found and their potential impact, but with no quantitative assessment, providing recommendations without grading the level of impact, leaving duty bearers to define a proper action plan.

This approach to HRIA is in line with voluntary and policy-based HRIA practice in the business sector. However, once HRIA becomes a legal tool – as suggested by the European Commission and the Council of Europe –, it is no longer merely a source of recommendations for better business policy. Future AI regulation will most likely bring specific legal obligations and sanctions for non-compliance in relation to risk assessment and management.

Analysis of potential impact will therefore become an element of regulatory compliance, with mandatory adoption of appropriate mitigation measures, and barriers in the event of high risk. A model that enables a graduation of risk can therefore facilitate compliance and reduce risks by preventing high-risk AI applications from being placed on the market.

With large-scale projects, such as smart cities, assessing each technological component using the proposed model and mitigating adverse effects is not sufficient. A more general overall analysis must be conducted in addition. Only an integrated assessment can consider the cumulative effect of a socio-technical system by measuring its broader impacts, including the consequences in terms of democratic participation and decision-making processes.

While the assessment of individual AI products/services might be carried out by the developing entity using the proposed model, large-scale multi-factor scenarios will require an additional layer: an integrated impact assessment conducted by external advisors. This integrated assessment, based on broader fieldwork, citizen engagement, and a co-design process, can evaluate the overall impact of an entire AI-based environment, in a way that is closer to traditional HRIA models.

In both cases, figures such as the human rights officer and tools like a HRIA management plan, containing action plans with timelines, responsibilities and indicators, can facilitate these processes,[251] including the possibility of extending them to the supply chain and all potentially affected groups of people.

Finally, the proposed model with its more formalised assessment can facilitate the accountability and monitoring of AI products and services during their lifecycle,[252] enabling changes in their impacts to be monitored through periodic reviews, audits, and progress reports on the implementation of the measures taken. It also makes it possible to incorporate more precise human rights indicators in internal reports and plans and make assessment results available to stakeholders clearly and understandably, facilitating their cooperation in a human rights-oriented approach to AI.

---

[250] See e.g. The Danish Institute for Human Rights, 'Human Rights Impact Assessment – Durex and Enfa value chains in Thailand' (2020) <https://www.humanrights.dk/publications/human-rights-impact-assessment-durex-enfa-value-chains-thailand> accessed 2 March 2021. But see Kendyl Salcito and Mark Wielga, 'Kayelekera HRIA Moitorig Summary' (2015) <http://nomogaia.org/wp-content/uploads/2015/10/KAYELEKERA-HRIA-MONITORING-SUMMARY-10-5-2015-Final.pdf> accessed 20 February 2021.

[251] See also Désirée Abrahams and Yann Wyss, 'Guide to Human Rights Impact Assessment and Management (HRIAM)' (The International Business Leaders Forum and the International Finance Corporation 2010) <https://d306pr3pise04h.cloudfront.net/docs/issues_doc%2Fhuman_rights%2FGuidetoHRIAM.pdf> accessed 26 October 2020.

[252] See also The Danish Institute for Human Rights, 'Guidance on HRIA of Digital Activities. Phase 4: Impact prevention, mitigation and remediation' (2020), 25-33 <https://www.humanrights.dk/sites/humanrights.dk/files/media/document/Phase%204_%20Impact%20prevention%2C%20mitigation%20and%20remediation_n.pdf> accessed 20 February 2021.